\documentclass[10pt,journal,compsoc]{IEEEtran}
\usepackage{graphicx}

\ifCLASSOPTIONcompsoc
  \usepackage[nocompress]{cite}
\else
  \usepackage{cite}
\fi
\usepackage{hyperref}       
\usepackage{url}            
\usepackage{epsfig}
\usepackage{graphicx}
\usepackage{booktabs}       
\usepackage{amsfonts}       
\usepackage{nicefrac}       
\usepackage{microtype}      
\usepackage{xcolor}         
\usepackage{amsmath}
\usepackage{amsfonts}
\usepackage{amssymb}
\usepackage{cleveref}
\usepackage{cancel}
\usepackage{amsthm}
\usepackage{wrapfig}
\usepackage{multirow}
\usepackage{subfig}
\usepackage{enumerate}
\usepackage{amsthm}
\usepackage{multirow}
\usepackage{wrapfig}
\usepackage{booktabs}
\usepackage{floatrow}
\usepackage{cancel}
\usepackage{makecell}
\usepackage[ruled,vlined]{algorithm2e}
\newtheorem{prop}{Proposition}
\newtheorem{duplicate}{Proposition}

\usepackage{enumitem}
\usepackage{listings}
\setlist[itemize]{leftmargin=*}
\newfloatcommand{capbtabbox}{table}[][\FBwidth]
\ifCLASSINFOpdf
\else
\fi

\def\ml{{\mathbf{l}}}
\def\mw{{\mathbf{w}}}
\def\mi{{\mathbf{i}}}
\def\mA{{\mathbf{A}}}
\def\mB{{\mathbf{B}}}
\def\mC{{\mathbf{C}}}

\def\mI{{\mathbf{I}}}
\def\mJ{{\mathbf{J}}}

\def\mP{{\mathbf{P}}}
\def\mQ{{\mathbf{Q}}}
\def\mR{{\mathbf{R}}}
\def\mS{{\mathbf{S}}}

\def\mU{{\mathbf{U}}}
\def\mV{{\mathbf{V}}}
\def\mW{{\mathbf{W}}}
\def\mX{{\mathbf{X}}}
\def\mY{{\mathbf{Y}}}

\begin{document}
%
\title{Orthogonal SVD Covariance Conditioning and Latent Disentanglement}
%
%
%
%
\author{Yue~Song,~\IEEEmembership{Member,~IEEE,}
        Nicu~Sebe,~\IEEEmembership{Senior Member,~IEEE,}
        Wei~Wang,~\IEEEmembership{Member,~IEEE}
\IEEEcompsocitemizethanks{\IEEEcompsocthanksitem Yue Song and Nicu Sebe are with Department
of Information Engineering and Computer Science, University of Trento, Trento 38123,
Italy. Wei Wang is with Beijing Jiaotong University, Beijing, China. Wei Wang is the corresponding author.\\
E-mail: \{yue.song, nicu.sebe\}@unitn.it, wei.wang@bjtu.edu.cn}
\thanks{Manuscript received April 19, 2005; revised August 26, 2015.}}

%
%

\markboth{IEEE TRANSACTIONS ON PATTERN ANALYSIS AND MACHINE INTELLIGENCE}%
{Shell \MakeLowercase{\textit{et al.}}: Bare Demo of IEEEtran.cls for Computer Society Journals}
%



\IEEEtitleabstractindextext{%
\begin{abstract}
Inserting an SVD meta-layer into neural networks is prone to make the covariance ill-conditioned, which could harm the model in the training stability and generalization abilities. In this paper, we systematically study how to improve the covariance conditioning by enforcing orthogonality to the Pre-SVD layer. Existing orthogonal treatments on the weights are first investigated. However, these techniques can improve the conditioning but would hurt the performance. To avoid such a side effect, we propose the Nearest Orthogonal Gradient (NOG) and Optimal Learning Rate (OLR). The effectiveness of our methods is validated in two applications: decorrelated Batch Normalization (BN) and Global Covariance Pooling (GCP). Extensive experiments on visual recognition demonstrate that our methods can simultaneously improve covariance conditioning and generalization. The combinations with orthogonal weight can further boost the performance. Moreover, we show that our orthogonality techniques can benefit generative models for better latent disentanglement through a series of experiments on various benchmarks. Code is available at: \href{https://github.com/KingJamesSong/OrthoImproveCond}{https://github.com/KingJamesSong/OrthoImproveCond}.
\end{abstract}


}

\maketitle

\IEEEdisplaynontitleabstractindextext

%
\IEEEpeerreviewmaketitle

\section{Introduction}

The Singular Value Decomposition (SVD) can factorize a matrix into orthogonal eigenbases and non-negative singular values, serving as an essential step for many matrix operations. Recently in computer vision and deep learning, many approaches integrated the SVD as a meta-layer in the neural networks to perform some differentiable spectral transformations, such as the matrix square root and inverse square root. The applications arise in a wide range of methods, including Global Covariance Pooling (GCP)~\cite{li2017second,song2021approximate,gao2021temporal}, decorrelated Batch Normalization (BN)~\cite{huang2018decorrelated,huang2021group,song2022fast}, Whitening an Coloring Transform (WCT) for universal style transfer~\cite{li2017universal,chiu2019understanding,wang2020diversified}, and Perspective-n-Point (PnP) problems~\cite{brachmann2017dsac,campbell2020solving,dang2020eigendecomposition}.

For the input feature map $\mX$ passed to the SVD meta-layer, one often first computes the covariance of the feature as $\mX\mX^{T}$. This can ensure that the covariance matrix is both symmetric and positive semi-definite, which does not involve any negative eigenvalues and leads to the identical left and right eigenvector matrices. However, it is observed that inserting the SVD layer into deep models would typically make the covariance very ill-conditioned~\cite{song2021approximate}, 
resulting in deleterious consequences on the stability and optimization of the training process. For a given covariance $\mA$, its conditioning is measured by the condition number:
\begin{equation}
    \kappa(\mA) = \sigma_{max}(\mA) \sigma_{min}^{-1}(\mA) 
\end{equation}
where $\sigma(\cdot)$ denotes the eigenvalue of the matrix. Mathematically speaking, the condition number measures how sensitive the SVD is to the errors of the input. Matrices with low condition numbers are considered \textbf{well-conditioned}, while matrices with high condition numbers are said to be \textbf{ill-conditioned}. Specific to neural networks, the ill-conditioned covariance matrices are harmful to the training process in several aspects, which we will analyze in detail later.

\begin{figure}[htbp]
    \centering
    \includegraphics[width=0.99\linewidth]{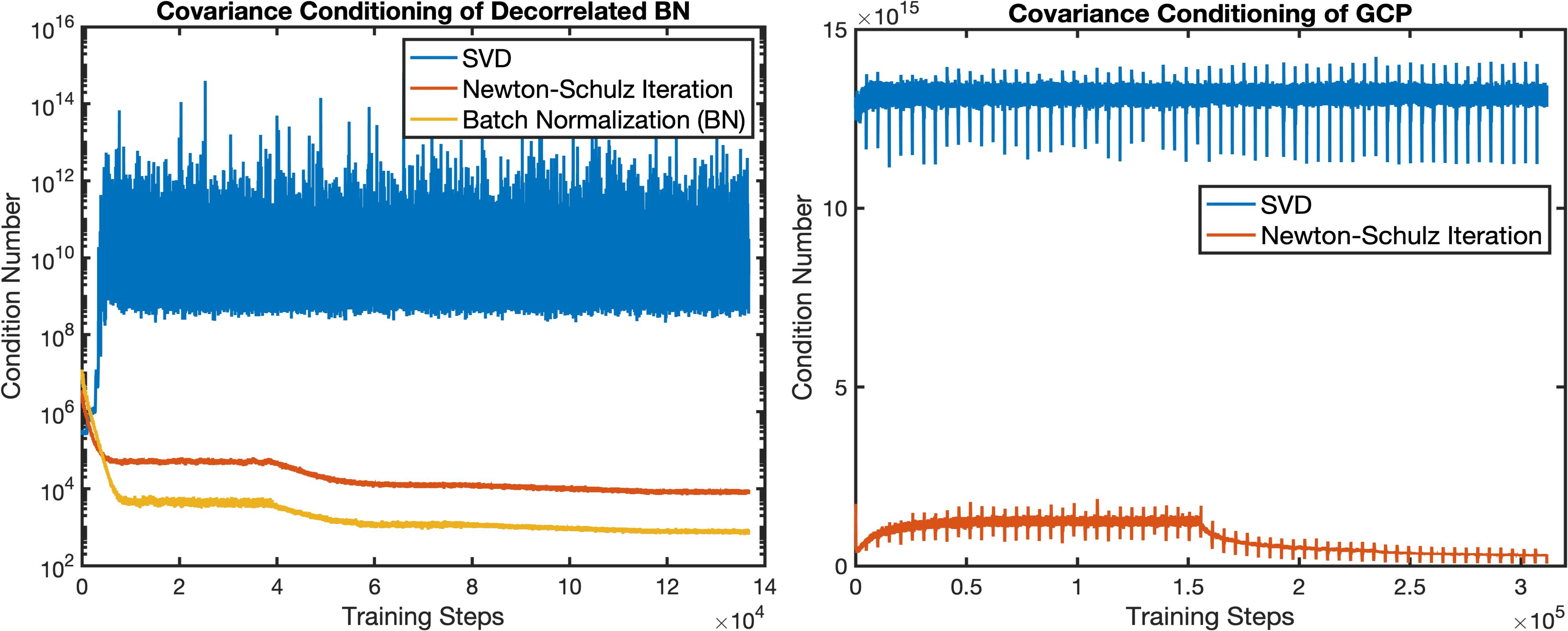}
    \caption{The covariance conditioning of the SVD meta-layer during the training process in the tasks of decorrelated BN (\emph{left}) and GCP (\emph{Right}). The decorrelated BN is based on ResNet-50 and CIFAR100, while ImageNet and ResNet-18 are used for the GCP.}
    \label{fig:cover_cond}
\end{figure}

This phenomenon was first observed in the GCP methods by~\cite{song2021approximate}, and we found that it generally extrapolates to other SVD-related tasks, such as decorrelated BN. Fig.~\ref{fig:cover_cond} depicts the covariance conditioning of these two tasks throughout the training. As can be seen, the integration of the SVD layer makes the generated covariance very ill-conditioned  (${\approx}1e12$ for decorrelated BN and ${\approx}1e16$ for GCP). By contrast, the conditioning of the approximate 
solver, \emph{i.e.,} Newton-Schulz iteration (NS iteration)~\cite{higham2008functions}, is about $1e5$ for decorrelated BN and is around $1e15$ for GCP, while the standard BN only has a condition number of $1e3$.


Ill-conditioned covariance matrices can harm the training of the network in both the forward pass (FP) and the backward pass (BP). For the FP, mainly the SVD solver is influenced in terms of stability and accuracy. Since the ill-conditioned covariance has many trivially-small eigenvalues, it is difficult for an SVD solver to accurately estimate them and large round-off errors are likely to be triggered, which might hurt the network performances. Moreover, the very imbalanced eigenvalue distribution can easily make the SVD solver fail to converge and cause the training failure~\cite{wang2021robust,song2021approximate}. For the BP, as pointed out in~\cite{lecun2012efficient,wiesler2011convergence,huang2018decorrelated}, the feature covariance is closely related to the Hessian matrix during the backpropagation. As the error curvature is given by the eigenvalues of the Hessian matrix~\cite{sutskever2013importance}, for the ill-conditioned Hessian, the Gradient Descent (GD) step would bounce back and forth in high curvature directions (large eigenvalues) and make slow progress in low curvature directions (small eigenvalues). As a consequence, the ill-conditioned covariance could cause slow convergence and oscillations in the optimization landscape. The generalization abilities of a deep model are thus harmed.

Due to the data-driven learning nature and the highly non-linear transform of deep neural networks, directly giving the analytical form of the covariance conditioning is intractable. Some simplifications have to be performed to ease the investigation. Since the covariance is generated and passed from the previous layer, the previous layer is likely to be the most relevant to the conditioning. Therefore, we naturally limit our focus to the Pre-SVD layer, \emph{i.e.,} the layer before the SVD layer. To further simplify the analysis, we study the Pre-SVD layer in two consecutive training steps, which can be considered as a mimic of the whole training process. Throughout the paper, we mainly investigate some meaningful manipulations on the weight, the gradient, and the learning rate of the Pre-SVD layer in two sequential training steps. \textit{Under our Pre-SVD layer simplifications, one promising direction to improve the conditioning is enforcing orthogonality on the weights.} Orthogonal weights have the norm-preserving property, which could improve the conditioning of the feature matrix. This technique has been widely studied in the literature of stable training and Lipschitz networks~\cite{mishkin2016all,wang2020orthogonal,singla2021skew}. We select some representative methods and validate their effectiveness in the task of decorrelated BN. Our experiment reveals that these orthogonal techniques can greatly improve the covariance conditioning, but could only bring marginal performance improvements and even slight degradation. \textit{This indicates that when the representation power of weight is limited, the improved conditioning does not necessarily lead to better performance. Orthogonalizing only the weight is thus insufficient to improve the generalization.} Instead of seeking orthogonality constraints on the weights, we propose our Nearest Orthogonal Gradient (NOG) and Optimal Learning Rate (OLR). These two techniques explore the orthogonality possibilities about the learning rate and the gradient. More specifically, our NOG modifies the gradient of the Pre-SVD layer into its nearest-orthogonal form and keeps the GD direction unchanged. On the other hand, the proposed OLR dynamically changes the learning rate of the Pre-SVD layer at each training step such that the updated weight is as close to an orthogonal matrix as possible. The experimental results demonstrate that the proposed two techniques not only significantly improve the covariance conditioning but also bring obvious improvements in the validation accuracy of both GCP and decorrelated BN. Moreover, when combined with the orthogonal weight treatments, the performance can have further improvements. 




Besides the application on differentiable SVD, we propose that our orthogonality techniques can be also used for unsupervised latent disentanglement of Generative Adversarial Networks (GANs)~\cite{goodfellow2014generative}. Recent works~\cite{zhu2021low,shen2021closed} revealed that the latent disentanglement of GANs is closely related to the gradient or weight of the first projector after the latent code. In particular, the eigenvectors of the gradient or weight can be viewed as closed-formed solutions of interpretable directions~\cite{shen2021closed}. This raises the need for enforcing orthogonal constraints on the projector. \textit{As shown in Fig.~\ref{fig:ortho_illu}, compared with non-orthogonal matrices, orthogonal matrices can lead to more disentangled representations and more precise attributes due to the property of equally-important eigenvectors.} Motivated by this observation, we propose to enforce our NOG and OLR as orthogonality constraints in generative models. Extensive experiments on various architectures and datasets demonstrate that our methods indeed improve the disentanglement ability of identifying semantic attributes and achieve state-of-the-art performance against other disentanglement approaches.



The main contributions are summarized below:
\begin{itemize}
    \item We systematically study the problem of how to improve the covariance conditioning of the SVD meta-layer. We propose our Pre-SVD layer simplification to investigate this problem from the perspective of orthogonal constraints. 
    \item We explore different techniques of orthogonal weights to improve the covariance conditioning. Our experiments reveal that these techniques could improve the conditioning but would harm the generalization abilities due to the limitation on the representation power of weight.
    \item We propose the nearest orthogonal gradient and optimal learning rate. The experiments on GCP and decorrelated BN demonstrate that these methods can attain better covariance conditioning and improved generalization. Their combinations with weight treatments can further boost the performance. 
    \item We show that our proposed orthogonality approaches can be applied on the GANs projector for improved latent disentanglement ability of discovering precise semantic attributes, which opens the way for new applications of orthogonality techniques.
\end{itemize}

This paper is an extension of the previous conference paper~\cite{song2022improving}. In~\cite{song2022improving}, we propose two orthogonality techniques and demonstrate that these methods can simultaneously improve the covaraince conditioning and generalization abilities of the SVD meta-layer. This journal extension motivates and proposes that these techniques can be also applied in generative models for better latent disentanglement. This point is validated through extensive experiments on various generative architectures and datasets. Moreover, we also investigate the probability of occurrence of our OLR throughout the training and show that the evaluation results agree well with our theoretical analysis. 

The rest of the paper is organized as follows: Sec.~\ref{sec:related} describes the related work in differentiable matrix decomposition, orthogonality applications, and unsupervised latent disentanglement. Sec.~\ref{sec:pre_svd_and_weight} introduces our Pre-SVD layer simplification and orthogonal weight treatments, and Sec.~\ref{sec:NOG_OLR} presents the proposed orthogonality techniques. Sec.~\ref{sec:ortho_latent} motivates why orthogonality can improve latent disentanglement. Sec.~\ref{sec:exp} provides experimental results and some in-depth analysis. Finally, Sec.~\ref{sec:conclusion} summarizes the conclusions.

\section{Related Work}
\label{sec:related}

\subsection{Differentiable Matrix Decomposition}

The differentiable matrix decomposition is widely used in neural networks as a spectral meta-layer. Ionescu~\emph{et al.}~\cite{ionescu2015matrix,ionescu2015training} first propose the theory of matrix back-propagation and laid a foundation for the follow-up research. In deep neural networks, the transformation of matrix square root and its inverse are often desired due to the appealing spectral property. Their applications cover a wide range of computer vision tasks~\cite{song2022fast,song2022fast2}. To avoid the huge time consumption of the SVD, some iterative methods are also developed to approximate the solution~\cite{higham2008functions,song2022fast,song2022fast2}. Recently Song~\emph{et al.}~\cite{song2022batch} propose a dedicated eigen-solver for improving the computation speed of batched matrices. In~\cite{huang2018decorrelated,chiu2019understanding,huang2019iterative,huang2020investigation,huang2021group,song2022fast}, the inverse square root is used in the ZCA whitening transform to whiten the feature map, which is also known as the decorrelated BN. The Global Covariance Pooling (GCP) models~\cite{li2017second,li2018towards,wang2020deep,xie2021so,song2021approximate,gao2021temporal,song2022eigenvalues} compute the matrix square root of the covariance as a spectral normalization, which achieves impressive performances on some recognition tasks, including large-scale visual classification~\cite{li2017second,song2021approximate,xie2021so,song2022fast}, fine-grained visual categorization~\cite{li2017second,li2018towards,song2022eigenvalues}, and video action recognition~\cite{gao2021temporal}. The Whitening and Coloring Transform (WCT), which uses both the matrix square root and inverse square root, is usually adopted in some image generation tasks such as neural style transfer~\cite{li2017universal,wang2020diversified}, image translation~\cite{ulyanov2017improved,cho2019image}, and domain adaptation~\cite{abramov2020keep,choi2021robustnet}. In the geometric vision problems, the differentiable SVD is usually applied to estimate the fundamental matrix and the camera pose~\cite{ranftl2018deep,dang2020eigendecomposition,campbell2020solving}. Besides the SVD-based factorization, differentiating Cholesky decomposition~\cite{murray2016differentiation} and some low-rank decomposition is used to approximate the attention mechanism~\cite{geng2020attention,xiong2021nystromformer,lu2021soft} or to learn the constrained representations~\cite{chan2015pcanet,yang2017admm}.  

\subsection{Orthogonality in Neural Network}

Orthogonal weights have the benefit of the norm-preserving property, \emph{i.e.,} the relation $||\mW\mA||_{\rm F}{=}||\mA||_{\rm F}$ holds for any orthogonal $\mW$. When it comes to deep neural networks, such a property can ensure that the signal stably propagates through deep networks without either exploding or vanishing gradients~\cite{bengio1994learning,glorot2010understanding}, which could speed up convergence and encourage robustness and generalization. In general, there are three ways to enforce orthogonality to a layer: orthogonal weight initialization~\cite{saxe2014exact,mishkin2016all,xiao2018dynamical}, orthogonal regularization~\cite{rodriguez2016regularizing,bansal2018can,qi2020deep,bansal2018can,wang2020orthogonal}, and explicit orthogonal weight via Carley transform or matrix exponential~\cite{maduranga2019complex,trockman2020orthogonalizing,singla2021skew}. Among these techniques, orthogonal regularization and orthogonal weight are most commonly used as they often bring some practical improvements in generalization. Since the covariance is closely related to the weight matrix of the Pre-SVD layer, enforcing the orthogonality constraint could help to improve the covariance conditioning of the SVD meta-layer. We will choose some representative methods and validate their impact in Sec.~\ref{sec:general_orthogonality}. 

Notice that the focus of existing literature is different from our work. The orthogonality constraints are often used to improve the Lipschitz constants of the neural network layers, which is expected to improve the visual quality in image generation~\cite{brock2018large,miyato2018spectral}, to allow for better adversarial robustness~\cite{tsuzuku2018lipschitz,singla2021skew}, and to improve generalization abilities~\cite{sedghi2018singular,wang2020orthogonal}. Our work is concerned with improving the covariance conditioning and generalization performance. Moreover, the orthogonality literature mainly investigates how to enforce orthogonality to weight matrices, whereas less attention is put on the gradient and learning rate. In Sec.~\ref{sec:NOG_OLR}, we will explore such possibilities and propose our solutions: nearest orthogonal gradient and optimal learning rate which is optimal in the sense that the updated weight is as close to an orthogonal matrix as possible. 

\subsection{Unsupervised Latent Disentanglement of GANs}
Interpreting latent spaces of GAN models in an unsupervised manner has received wide attention recently~\cite{bau2019gan,jahanian2020steerability,voynov2020unsupervised,tzelepis2021warpedganspace}. This can help to identify semantic attributes of the image and to have precise control of the generation process, which could benefit both local and global image editing tasks~\cite{shen2020interpreting,zhu2021low}. Voynov~\emph{et al.}~\cite{voynov2020unsupervised} proposed to jointly learn a set of directions and an extra classifier such that the interpretable directions can be recognized. In~\cite{harkonen2020ganspace}, the authors proposed to perform PCA on the sampled data to capture the interpretable directions. More recently, Shen~\emph{et al.}~\cite{shen2021closed} and Zhu~\emph{et al.}~\cite{zhu2021low} pointed out that the semantic attributes are characterized by the eigenvectors of the weight or gradient of the first projector after the latent code. Motivated by this observation, we propose to enforce our orthogonality techniques to the gradient and weight matrices. 

Besides our orthogonality techniques, a few works have applied implicit orthogonality into the training process of GANs to attain more disentangled representations~\cite{peebles2020hessian,zhu2020learning,he2021eigengan,wei2021orthogonal}. In~\cite{peebles2020hessian,wei2021orthogonal}, the authors proposed to add orthogonal Hessian/Jacobian penalty to encourage disentanglement. He~\emph{et al.}~\cite{he2021eigengan} designed a dedicated GAN architecture where multi-level latent codes and orthogonal weight constraints are applied. Different from previous approaches, our orthogonality treatments do not rely on any implicit regularization. Instead, our NOG explicitly maps the original gradient into the nearest-orthogonal form, while our OLR keeps the updated weight in the closest form to orthogonal matrices. 

\section{Pre-SVD Layer and Weight Treatments}
\label{sec:pre_svd_and_weight}

In this section, we first motivate our simplification of the Pre-SVD layer, and then validate the efficacy of some representative weight treatments.

\subsection{Pre-SVD Layer Simplification}
\label{sec:pre_svd}
The neural network consists of a sequential of non-linear layers where the learning of each layer is data-driven. Stacking these layers leads to a highly non-linear and complex transform, which makes directly analyzing the covariance conditioning intractable. To solve this issue, we have to perform some simplifications.  

Our simplifications involve limiting the analysis only to the layer previous to the SVD layer (which we dub as the Pre-SVD layer) in two consecutive training steps. The Pre-SVD layer directly determines the conditioning of the generated covariance, while the two successive training steps are a mimic of the whole training process. The idea is to simplify the complex transform by analyzing the sub-model (two layers) and the sub-training (two steps), which can be considered as an "abstract representation" of the deep model and its complete training.

Let $\mW$ denote the weight matrix of the Pre-SVD layer. Then for the input $\mX_{l}$ passed to the layer, we have:
\begin{equation}
    \mX_{l+1} = \mW\mX_{l} + \mathbf{b}
\end{equation}
where $\mX_{l+1}$ is the feature passed to the SVD layer, and $\mathbf{b}$ is the bias vector. 
Since the bias $\mathbf{b}$ has a little influence here, we can sufficiently omit it for simplicity. The covariance in this step is computed as $\mW\mX_{l}\mX_{l}^{T}\mW^{T}$.
After the BP, the weight matrix is updated as $\mathbf{W}{-}{\eta}\frac{\partial l}{\partial \mathbf{W}}$ where $\eta$ denotes the learning rate of the layer. Let $\mY_{l}$ denote the passed-in feature of the next training step. Then the covariance is calculated as:
\begin{equation}
\begin{aligned}
    \mC &= \Big( (\mathbf{W}-\eta\frac{\partial l}{\partial \mathbf{W}})\cdot\mY_{l} \Big)\Big( (\mathbf{W}-\eta\frac{\partial l}{\partial \mathbf{W}})\cdot\mY_{l} \Big)^{T}\\
    &=\begin{gathered}
         \mW\mY_{l}\mY_{l}^{T}\mW^{T} {-} \eta\frac{\partial l}{\partial \mathbf{W}}\mY_{l}\mY_{l}^{T}\mW^{T}\\ {-} \eta\mW\mY_{l}\mY_{l}^{T}(\frac{\partial l}{\partial \mathbf{W}})^{T} {+} \eta^{2}\frac{\partial l}{\partial \mathbf{W}}\mY_{l}\mY_{l}^{T}(\frac{\partial l}{\partial \mathbf{W}})^{T}
    \end{gathered}
    \label{eq:problem}
\end{aligned}
\end{equation}
where $\mC$ denotes the generated covariance of the second step. Now the problem becomes how to stop the new covariance $\mC$ from becoming worse-conditioned than $\mW\mX_{l}\mX_{l}^{T}\mW^{T}$. In eq.~\eqref{eq:problem}, three variables could influence the conditioning: the weight $\mW$, the gradient of the last step $\frac{\partial l}{\partial \mW}$, and the learning rate $\eta$ of this layer. Among them, the weight $\mW$ seems to be the most important as it contributes to three terms of eq.~\eqref{eq:problem}. Moreover, the first term $\mW\mY_{l}\mY_{l}^{T}\mW^{T}$ computed by $\mW$ is not attenuated by $\eta $ or $\eta^2$ like the other terms. Therefore, it is natural to first consider manipulating $\mW$ such that the conditioning of $\mC$ could be improved. 

\subsection{General Treatments on Weights}
\label{sec:general_orthogonality}

In the literature of enforcing orthogonality to the neural network, there are several techniques to improve the conditioning of the weight $\mW$. Now we introduce some representatives methods and validate their impacts. 

\subsubsection{Spectral Normalization (SN)} In~\cite{miyato2018spectral}, the authors propose a normalization method to stabilize the training of generative models~\cite{goodfellow2014generative} by dividing the weight matrix with its largest eigenvalue. The process is defined as:
\begin{equation}
    \mW / \sigma_{max}(\mW)
\end{equation}
Such a normalization can ensure that the spectral radius of $\mW$ is always $1$, \emph{i.e.,} $\sigma_{max}(\mW){=}1$. This could help to reduce the conditioning of the covariance since we have $\sigma_{max}(\mW\mY_{l}){=}\sigma_{max}(\mY_{l})$ after the spectral normalization.

\subsubsection{Orthogonal Loss (OL)} Besides limiting the spectral radius of $\mW$, enforcing orthogonality constraint could also improve the covariance conditioning. As orthogonal matrices are norm-preserving (\emph{i.e.,} $||\mW\mY_{l}||_{\rm F}{=}||\mW||_{\rm F}$), lots of methods have been proposed to encourage orthogonality on weight matrices for more stable training and better signal-preserving property~\cite{pascanu2013difficulty,bansal2018can,wang2020orthogonal,trockman2020orthogonalizing,singla2021skew}. One common technique is to apply \emph{soft} orthogonality~\cite{wang2020orthogonal} by the following regularization:
\begin{equation}
    l=||\mW\mW^{T}-\mI||_{\rm F}
\end{equation}
This extra loss is added in the optimization objective to encourage more orthogonal weight matrices. However, since the constraint is achieved by regularization, the weight matrix is not exactly orthogonal at each training step.  

\subsubsection{Orthogonal Weights (OW)} Instead of applying \emph{soft} orthogonality by regularization, some methods can explicitly enforce \emph{hard} orthogonality to the weight matrices~\cite{trockman2020orthogonalizing,singla2021skew}. The technique of~\cite{singla2021skew} is built on the mathematical property: for any skew-symmetric matrix, its matrix exponential is an orthogonal matrix.
\begin{equation}
    \exp(\mW-\mW^{T})\exp(\mW-\mW^{T})^{T}=\mI
\end{equation}
where the operation of $\mW{-}\mW^{T}$ is to make the matrix skew-symmetric, \emph{i.e.,} the relation $\mW{-}\mW^{T}{=}-(\mW{-}\mW^{T})^{T}$ always holds. Then $\exp(\mW{-}\mW^{T})$ is used as the weight. This technique explicitly constructs the weight as an orthogonal matrix. The orthogonal constraint is thus always satisfied during the training.

\begin{figure}\CenterFloatBoxes
\begin{floatrow}
\ffigbox{%
  \includegraphics[width=0.99\linewidth]{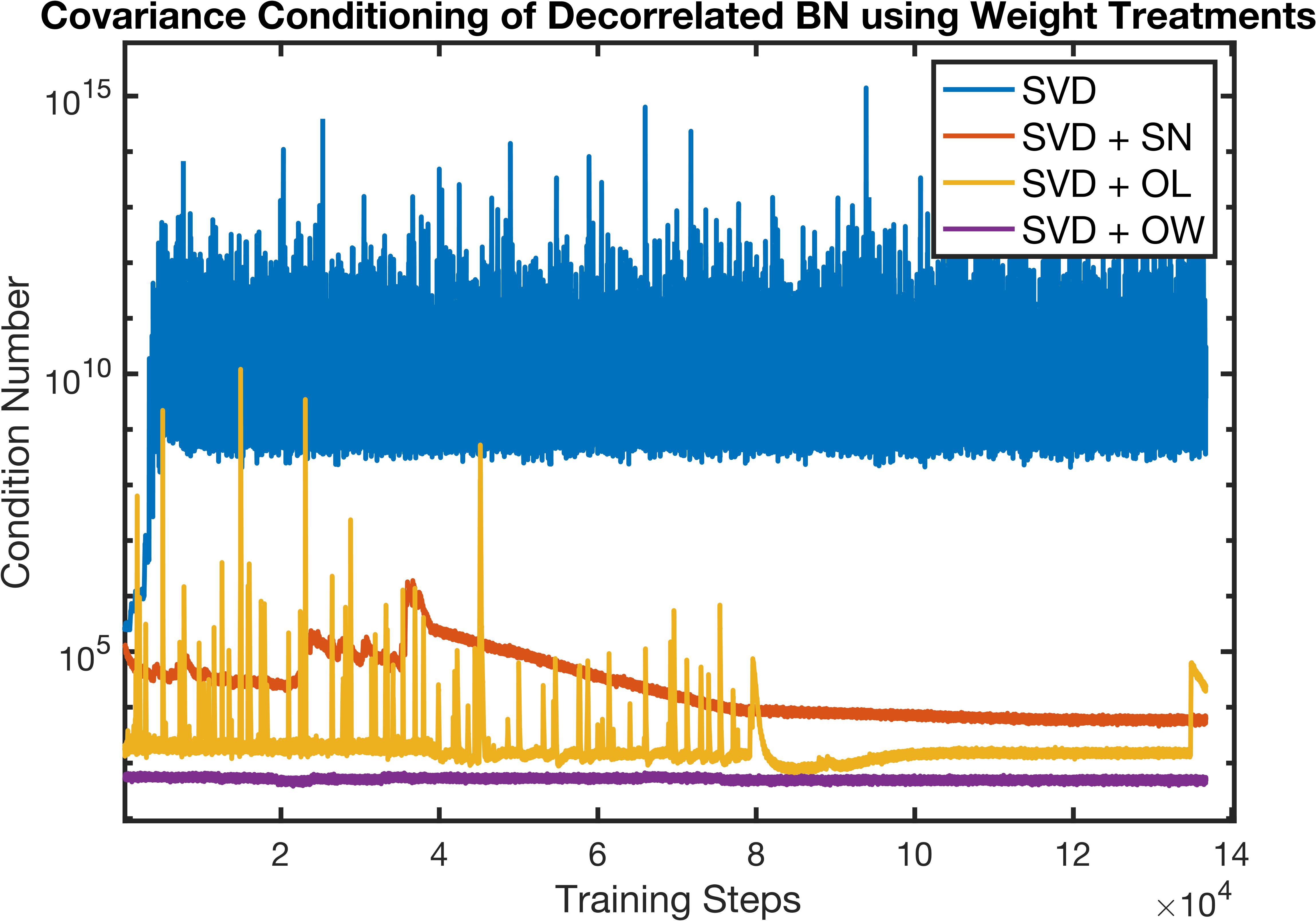}
}{%
  \caption{Covariance conditioning during the training process. All weight treatments can improve conditioning.}%
  \label{fig:ortho_weight}
}
\capbtabbox{%
\resizebox{0.99\linewidth}{!}{
  \begin{tabular}{r|c|c} \toprule
  Methods & mean$\pm$std & min \\ \hline
  SVD & 19.99$\pm$0.16 &19.80 \\ \hline
  SVD + SN & 19.94$\pm$0.33 &19.60 \\
  SVD + OL & \textbf{19.73$\pm$0.28} & \textbf{19.54} \\
  SVD + OW & 20.06$\pm$0.17 &19.94 \\
  \hline\hline
    NS iteration &19.45$\pm$0.33&19.01\\
  \bottomrule
  \end{tabular}
}
}
{
  \caption{Performance of different weight treatments on ResNet-50 and CIFAR100 based on $10$ runs.}%
  \label{tab:ortho_weight}
}
\end{floatrow}
\end{figure}

We apply the above three techniques in the experiment of decorrelated BN. Fig.~\ref{fig:ortho_weight} displays the covariance conditioning throughout the training, and Table~\ref{tab:ortho_weight} presents the corresponding validation errors. As can be seen, all of these techniques attain much better conditioning, but the performance improvements are not encouraging. The SN reduces the conditioning to around $10^{5}$, while the validation error marginally improves. The \emph{soft} orthogonality by the OL brings slight improvement on the performance despite some variations in the conditioning. The conditioning variations occur because the orthogonality constraint by regularization is not strictly enforced. Among the weight treatments, the \emph{hard} orthogonality by the OW achieves the best covariance conditioning, continuously maintaining the condition number around $10^{3}$ throughout the training. However, the OW slightly hurts the validation error. This implies that better covariance conditioning does not necessarily correspond to the improved performance, and orthogonalizing only the weight cannot improve the generalization. \textit{We conjecture that enforcing strict orthogonality only on the weight might limit its representation power.} Nonetheless, as will be discussed in Sec.~\ref{sec:NOG}, the side effect can be canceled when we simultaneously orthogonalize the gradient.

\section{Nearest Orthogonal Gradient and Optimal Learning Rate}
\label{sec:NOG_OLR}
This section introduces our proposed techniques on modifying the gradient and learning rate of the Pre-SVD layer. The combinations with weight treatments are also discussed.

\subsection{Nearest Orthogonal Gradient (NOG)}
\label{sec:NOG} 

As discussed in Sec.~\ref{sec:pre_svd}, the covariance conditioning is also influenced by the gradient $\frac{\partial l}{\partial \mW}$. However, existing literature mainly focuses on orthogonalizing the weights. To make the gradient also orthogonal, we propose to find the nearest-orthogonal gradient of the Pre-SVD layer. Different matrix nearness problems have been studied in~\cite{higham1988matrix}, and the nearest-orthogonal problem is defined as:
\begin{equation}
    \min_{\mR} ||\frac{\partial l}{\partial \mW}-\mR ||_{\rm F}\ subject\ to\ \mR\mR^{T}=\mI
\end{equation}
where $\mR$ is the seeking solution. To obtain such an orthogonal matrix, we can construct the error function as:
\begin{equation}
    e(\mR) = Tr\Big((\frac{\partial l}{\partial \mW}-\mR)^{T}(\frac{\partial l}{\partial \mW}-\mR)\Big) + Tr\Big(\mathbf{\Sigma} \mR^{T}\mR -\mI \Big) 
\end{equation}
where $Tr(\cdot)$ is the trace measure, and $\mathbf{\Sigma}$ denotes the symmetric matrix Lagrange multiplier. The closed-form solution is given by:
\begin{equation}
    \mR = \frac{\partial l}{\partial \mW} \Big(( \frac{\partial l}{\partial \mW})^{T} \frac{\partial l}{\partial \mW}\Big)^{-\frac{1}{2}}
\end{equation}
The detailed derivation is given in the supplementary material. If we have the SVD of the gradient ($\mU\mS\mV^{T}{=}\frac{\partial l}{\partial \mW}$), the solution can be further simplified as:
\begin{equation}
    \mR = \mU\mS\mV^{T} (\mV\mS^{-1}\mV^{T})=\mU\mV^{T} 
\end{equation}
As indicated above, the nearest orthogonal gradient is achieved by setting the singular value matrix to the identity matrix, \emph{i.e.,} setting $\mS$ to $\mI$. Notice that only the gradient of Pre-SVD layer is changed, while that of the other layers is not modified. Our proposed NOG can bring several practical benefits. 

\subsubsection{Orthogonal Constraint and Optimal Conditioning} The orthogonal constraint is exactly enforced on the gradient as we have $(\mU\mV^{T})^{T}\mU\mV^{T}{=}\mI$. Since we explicitly set all the singular values to $1$, the optimal conditioning is also achieved, \emph{i.e.,} $\kappa(\frac{\partial l}{\partial \mW}){=}1$. This could help to improve the conditioning. 


\subsubsection{Keeping Gradient Descent Direction Unchanged} In the high-dimensional optimization landscape, the many curvature directions (GD directions) are characterized by the eigenvectors of gradient ($\mU$ and $\mV$). Although our modification changes the gradient, the eigenvectors and the GD directions are untouched. In other words, our NOG only adjusts the step size in each GD direction. This indicates that the modified gradients will not harm performance. 

\subsubsection{Combination with Weight Treatments} 
\label{sec:ol_nog_combination}
Our orthogonal gradient and the previous weight treatments are complementary. They can be jointly used to simultaneously orthogonalize the gradient and weight. In the following, we will validate their joint impact on the conditioning and performance.


\begin{figure}\CenterFloatBoxes
\begin{floatrow}
\ffigbox{%
  \includegraphics[width=0.99\linewidth]{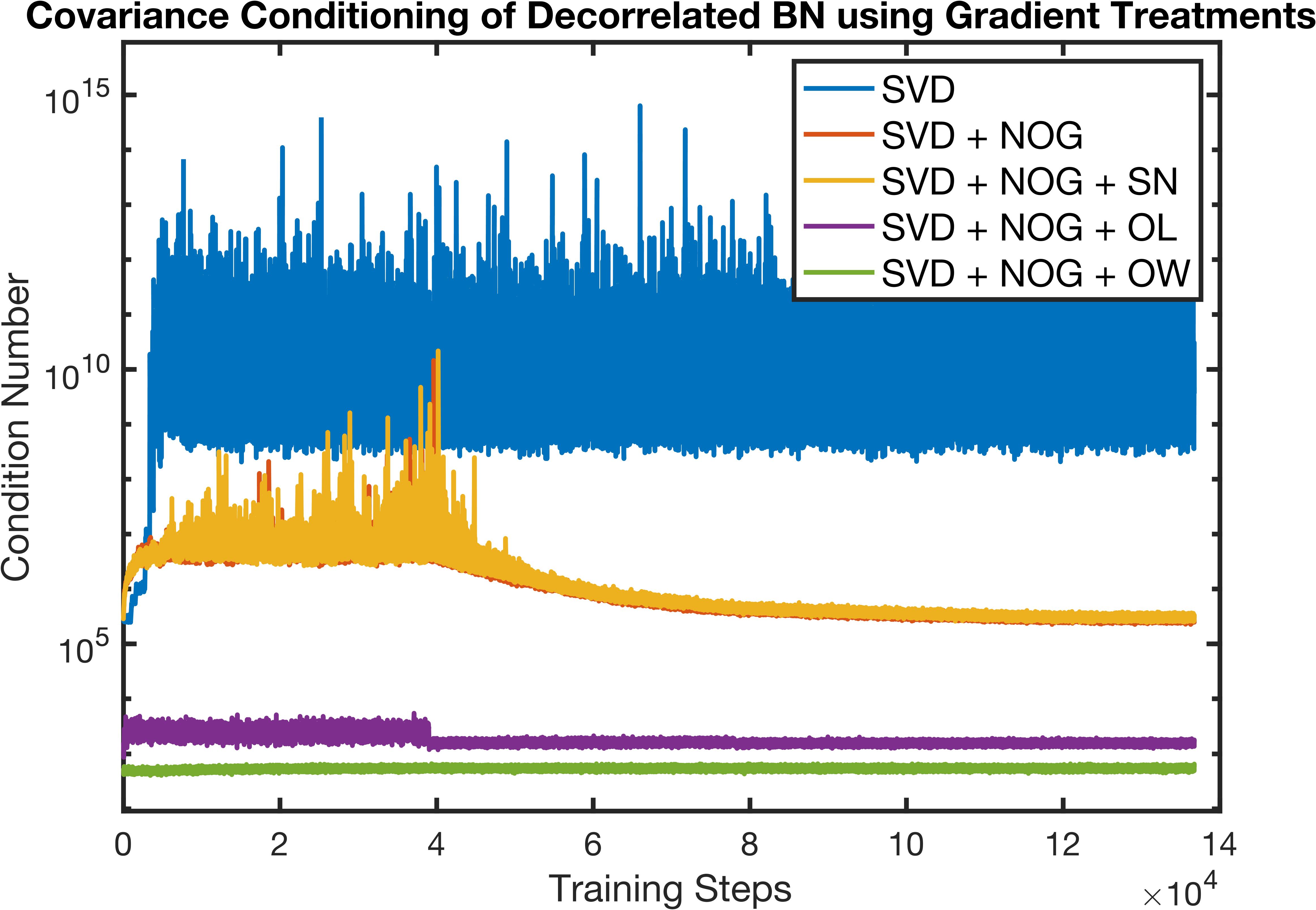}
}{%
  \caption{Covariance conditioning during the training process using orthogonal gradient and weight treatments.}%
  \label{fig:gradient}
}
\capbtabbox{%
\resizebox{0.99\linewidth}{!}{
  \begin{tabular}{r|c|c} \toprule
  Methods & mean$\pm$std & min \\ \hline
  SVD & 19.99$\pm$0.16 &19.80 \\ 
  SVD + NOG & 19.43$\pm$0.24 &19.15\\\hline
  SVD + NOG + SN & 19.43$\pm$0.21 &19.20 \\
  SVD + NOG + OL & 20.14$\pm$0.39 &19.54 \\
  SVD + NOG + OW & \textbf{19.22$\pm$0.28}&\textbf{18.90} \\
  \hline\hline
    NS iteration &19.45$\pm$0.33&19.01\\
  \bottomrule
  \end{tabular}
}
}{%
  \caption{Performance of gradient treatments on ResNet-50 and CIFAR100. Each result is based on $10$ runs.}%
  \label{tab:gradient}
}
\end{floatrow}
\end{figure}

Fig.~\ref{fig:gradient} and Table~\ref{tab:gradient} present the covariance conditioning of decorrelated BN and the corresponding validation errors, respectively. As we can observe, solely using the proposed NOG can largely improve the covariance conditioning, decreasing the condition number from $10^{12}$ to $10^6$. Though this improvement is not as significant as the orthogonal constraints (\emph{e.g.,} OL and OW), our NOG can benefit more the generalization abilities, leading to the improvement of validation error by $0.6\%$. Combining the SN with our NOG does not lead to obvious improvements in either the conditioning or validation errors, whereas the joint use of NOG and OL harms the network performances. This is because the orthogonality constraint by loss might not be enforced under the gradient manipulation. When our NOG is combined with the OW, the side effect of using only OW is eliminated and the performance is further boosted by $0.3\%$. This phenomenon demonstrates that when the gradient is orthogonal, applying the orthogonality constraint to the weight could also be beneficial to the generalization.


\subsection{Optimal Learning Rate (OLR)}
So far, we only consider orthogonalizing $\mW$ and $\frac{\partial l}{\partial \mW}$ separately, but 
how to jointly optimize $\mW{-}{\eta}\frac{\partial l}{\partial \mW}$ has not been studied yet. Actually, it is desired to choose an appropriate learning rate $\eta$ such that the updated weight is close to an orthogonal matrix. To this end, we need to achieve the following objective:
\begin{equation}
   \min_{\eta} ||(\mW-{\eta}\frac{\partial l}{\partial \mW})(\mW-{\eta}\frac{\partial l}{\partial \mW})^{T}-\mI||_{\rm F}
\end{equation}
This optimization problem can be more easily solved in the vector form. Let $\mathbf{w}$, $\mi$, and $\mathbf{l}$ denote the vectorized $\mW$, $\mI$, and $\frac{\partial l}{\partial \mW}$, respectively. Then we construct the error function as:
\begin{equation}
   e(\eta) = \Big((\mathbf{w}-\eta\mathbf{l})^{T}(\mathbf{w}-\eta\mathbf{l})-\mathbf{i}\Big)^{T}\Big((\mathbf{w}-\eta\mathbf{l})^{T}(\mathbf{w}-\eta\mathbf{l})-\mathbf{i}\Big)
\end{equation}
Expanding and differentiating the equation w.r.t. $\eta$ lead to:
\begin{equation}
\begin{gathered}
    \frac{d e(\eta)}{d \eta} \approx -4\mw\mw^{T}\ml^{T}\mw + 4\eta\mw\mw^{T}\ml^{T}\ml + 8\eta\ml^{T}\mw\ml^{T}\mw =0\\
   \eta^{\star} \approx \frac{\mw^{T}\mw\ml^{T}\mw}{\mw^{T}\mw\ml^{T}\ml+2\ml^{T}\mw\ml^{T}\mw}
   \label{optimal_lr}
\end{gathered}
\end{equation}
where some higher-order terms are neglected. The detailed derivation is given in the supplementary material. Though the proposed OLR yields the updated weight nearest to an orthogonal matrix theoretically, the value of $\eta^{\star}$ is unbounded for arbitrary $\mw$ and $\ml$. Directly using $\eta^{\star}$ might cause unstable training. To avoid this issue, we propose to use the OLR only when its value is smaller than the learning rate of other layers. Let $lr$ denote the learning rate of the other layers. The switch process can be defined as:
\begin{equation}
    \eta =\begin{cases}
    \eta^{\star} & if\ \eta^{\star}<lr\\
    lr & otherwise
    \end{cases}
\end{equation}

\subsubsection{Combination with Weight/Gradient Treatments} When either the weight or the gradient is orthogonal, our OLR needs to be carefully used. When only $\mW$ is orthogonal, $\mw^{T}\mw$ is a small constant and it is very likely to have $\mw^{T}\mw{\ll}\ml^{T}\mw$. Consequently, we have $\mw^{T}\mw\ml^{T}\mw{\ll}\ml^{T}\mw\ml^{T}\mw$ and $\eta^{\star}$ will attenuate to zero. Similarly for orthogonal gradient, we have $\mw^{T}\mw\ml^{T}\mw{\ll}\ml^{T}\mw\ml^{T}\ml$ and this will cause $\eta^{\star}$ close to zero. Therefore, the proposed OLR cannot work when either the weight or gradient is orthogonal. Nonetheless, we note that if both $\mW$ and $\frac{\partial l}{\partial \mW}$ are orthogonal, our $\eta^{\star}$ is bounded. Specifically, we have:

\begin{prop}
 When both $\mW$ and $\frac{\partial l}{\partial \mW}$ are orthogonal, $\eta^{\star}$ is both upper and lower bounded. The upper bound is $\frac{N^2}{N^2 + 2}$ and the lower bound is $\frac{1}{N^{2}+2}$ where $N$ denotes the row dimension of $\mW$.
\end{prop}

We give the detailed proof in the supplementary material. Obviously, the upper bound of $\eta^{\star}$ is smaller than $1$. For the lower bound, since the row dimension of $N$ is often large (\emph{e.g.,} $64$), the lower bound of $\eta^{\star}$ can be according very small (\emph{e.g.,} $2e{-}4$). This indicates that our proposed OLR could also give a small learning rate even in the later stage of the training process.



In summary, the optimal learning rate is set such that the updated weight is optimal in the sense that it become as close to an orthogonal matrix as possible. In particular, it is suitable when both the gradient and weight are orthogonal.




\begin{figure}\CenterFloatBoxes
\begin{floatrow}
\ffigbox{%
  \includegraphics[width=0.99\linewidth]{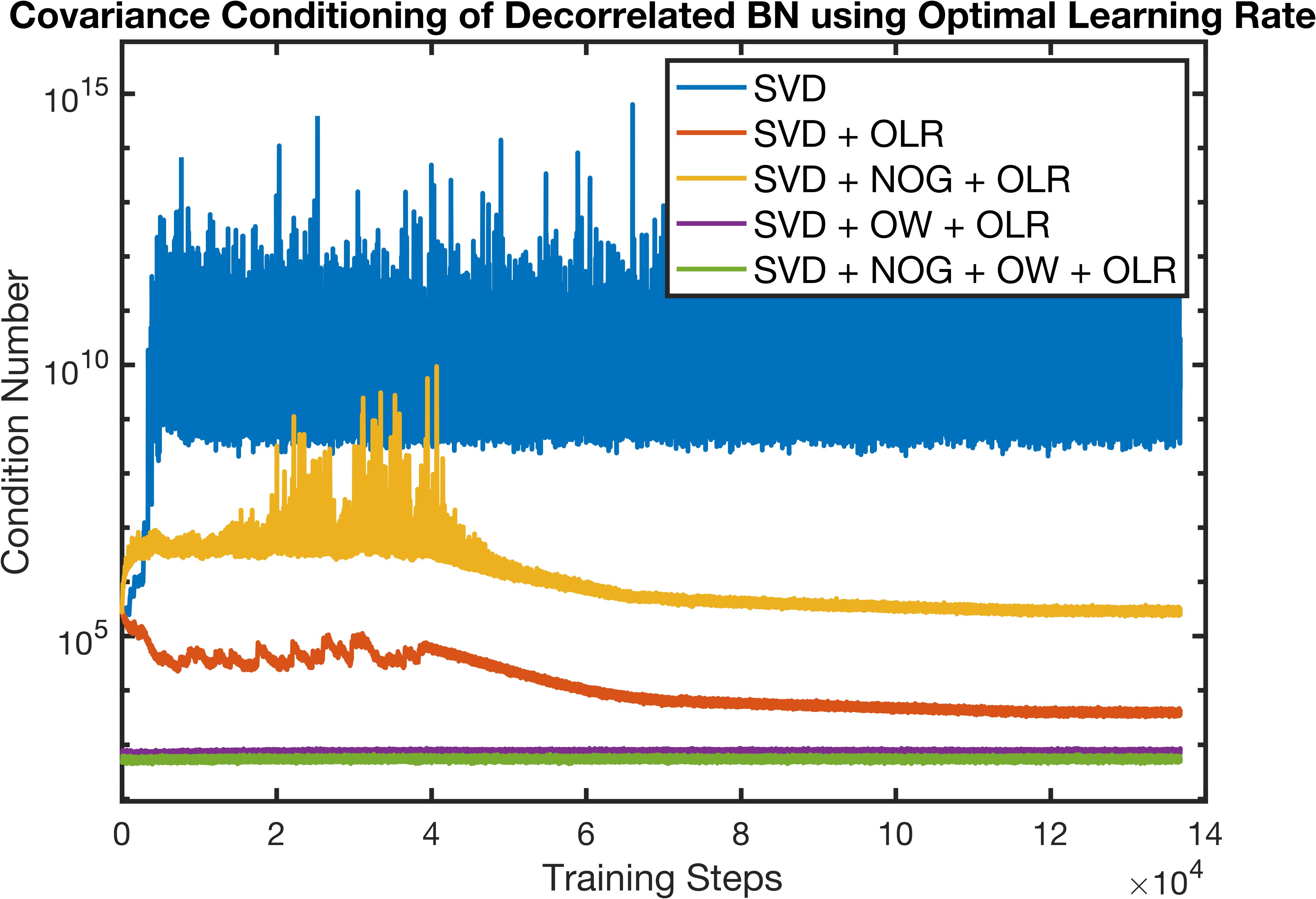}
}{%
  \caption{Covariance conditioning during the training process using optimal learning rate and hybrid treatments.}%
  \label{fig:olr}
}
\capbtabbox[\Xhsize]{%
\resizebox{0.99\linewidth}{!}{
  \begin{tabular}{r|c|c} \toprule
  Methods & mean$\pm$std & min \\ \hline
  SVD & 19.99$\pm$0.16 &19.80 \\ 
  SVD + OLR & 19.50$\pm$0.39 &18.95 \\\hline
  SVD + NOG + OLR & 19.77$\pm$0.27  &19.36 \\
  SVD + OW + OLR  & 20.61$\pm$0.22  &20.43 \\
  \makecell[c]{SVD + NOG \\+ OW +OLR} & \textbf{19.05$\pm$0.31}&\textbf{18.77} \\
  \hline\hline
    NS iteration &19.45$\pm$0.33&19.01\\
  \bottomrule
  \end{tabular}
}
}{%
  \caption{Performance of optimal learning rate on ResNet-50 and CIFAR100 based on $10$ runs.}
  \label{tab:olr}
}
\end{floatrow}
\end{figure}

We give the covariance conditioning and the validation errors in Fig.~\ref{fig:olr} and in Table~\ref{tab:olr}, respectively. Our proposed OLR significantly reduces the condition number to  $10^{4}$ and improves the validation error by $0.5\%$. When combined with either orthogonal weight or orthogonal gradient, there is a slight degradation on the validation errors. This meets our expectation as $\eta^{\star}$ would attenuate to zero in both cases. However, when both $\mW$ and $\frac{\partial l}{\partial \mW}$ are orthogonal, jointly using our OLR achieves the best performance, outperforming only OLR by $0.5\%$ and beating OW$+$NOG by $0.2\%$. This observation confirms that the proposed OLR works well for simultaneously orthogonal $\mW$ and $\frac{\partial l}{\partial \mW}$.

\section{Orthogonality for Unsupervised Latent Disentanglement}
\label{sec:ortho_latent}
In this section, we motivate why orthogonal treatments (orthogonal weight or gradient) would help in unsupervised latent disentanglement of GANs. 

\subsection{Image Manipulation in Latent Space of GANs}

The latent space of GANs encodes rich semantics information, which can be used for image editing via vector arithmetic property~\cite{radford2015unsupervised}. Consider a generator $G(\cdot)$ and the latent code $\mathbf{z}{\in}\mathbb{R}^{d}$. The image manipulation is achieved by finding a semantically meaningful direction $\mathbf{n}$ such that
\begin{equation}
    \texttt{edit}(G(\mathbf{z}))=G(\mathbf{z}+\alpha\mathbf{n}) 
    \label{eq:g}
\end{equation}
where $\texttt{edit}(\cdot)$ denotes the image editing process, and $\alpha$ represents the perturbation strength. That being said, moving the latent code $\mathbf{z}$ along with the interpretable direction $\mathbf{n}$ should change the targeting semantic concept of the image. Since the generator $G(\cdot)$ is highly non-linear and complex, directly analyzing $G(\mathbf{z}+\alpha\mathbf{n})$ is intractable. To avoid this issue, existing approaches propose to simplify the analysis by considering only the first projector matrix $G_{1}(\cdot)$ or performing local Taylor expansion~\cite{shen2021closed,zhu2021low,zhu2022region,balakrishnan2022rayleigh}. 

\noindent\textbf{Eigenvector of the first projector.} In SeFa~\cite{shen2021closed}, the authors propose to seek for interpretable directions from the eigenvector of the first projector matrix. Specifically, they consider the affine transformation of the layer as:
\begin{equation}
    G_{1}(\mathbf{z}+\alpha\mathbf{n}) = \mathbf{A}\mathbf{z} + \mathbf{b} + \alpha\mathbf{A}\mathbf{n} = G_{1}(\mathbf{z})  + \alpha\mathbf{A}\mathbf{n}
    \label{eq:g_1}
\end{equation}
where $\mathbf{A}$ is the weight matrix. Intuitively, a meaningful direction should lead to large variations of the generated image. So the problem can be cast into an optimization problem as:
\begin{equation}
    \mathbf{n}^{\star} = \arg\max ||\mathbf{A}\mathbf{n}||^{2}
    \label{eq:argmax_n}
\end{equation}
All the possible closed-form solution correspond to the eigenvector of $\mathbf{A}^{T}\mathbf{A}$. The top-$k$ eigenvectors are thus selected as the interpretable directions for image manipulation. 

\noindent\textbf{Eigenvector of the Jacobian.} LowRankGAN~\cite{zhu2021low} proposes to linearly approximate $G(\mathbf{z}+\alpha\mathbf{n})$ by the Taylor expansion as:
\begin{equation}
    G(\mathbf{z}+\alpha\mathbf{n}) \approx G(\mathbf{z}) + \alpha\mathbf{J}_{\mathbf{z}}\mathbf{n}
    \label{eq:g_jacob}
\end{equation}
where $\mathbf{J}_{\mathbf{z}}$ is the Jacobian matrix w.r.t. the latent code $\mathbf{z}$. Similarly to the deduction of eq.~\eqref{eq:argmax_n}, the closed-form solution is given by the eigenvector of $\mathbf{J}_{\mathbf{z}}^{T}\mathbf{J}_{\mathbf{z}}$. 

The above two formulations illustrate how the weight and gradient matrices are related with the interpretable direction discovery. Currently, most GAN models do not enforce orthogonality to their architectures. Now we turn to explaining the concrete benefit of introducing orthogonality to the latent disentanglement.


\subsection{Usefulness of Orthogonality}

\begin{figure}[htbp]
    \centering
    \includegraphics[width=0.99\linewidth]{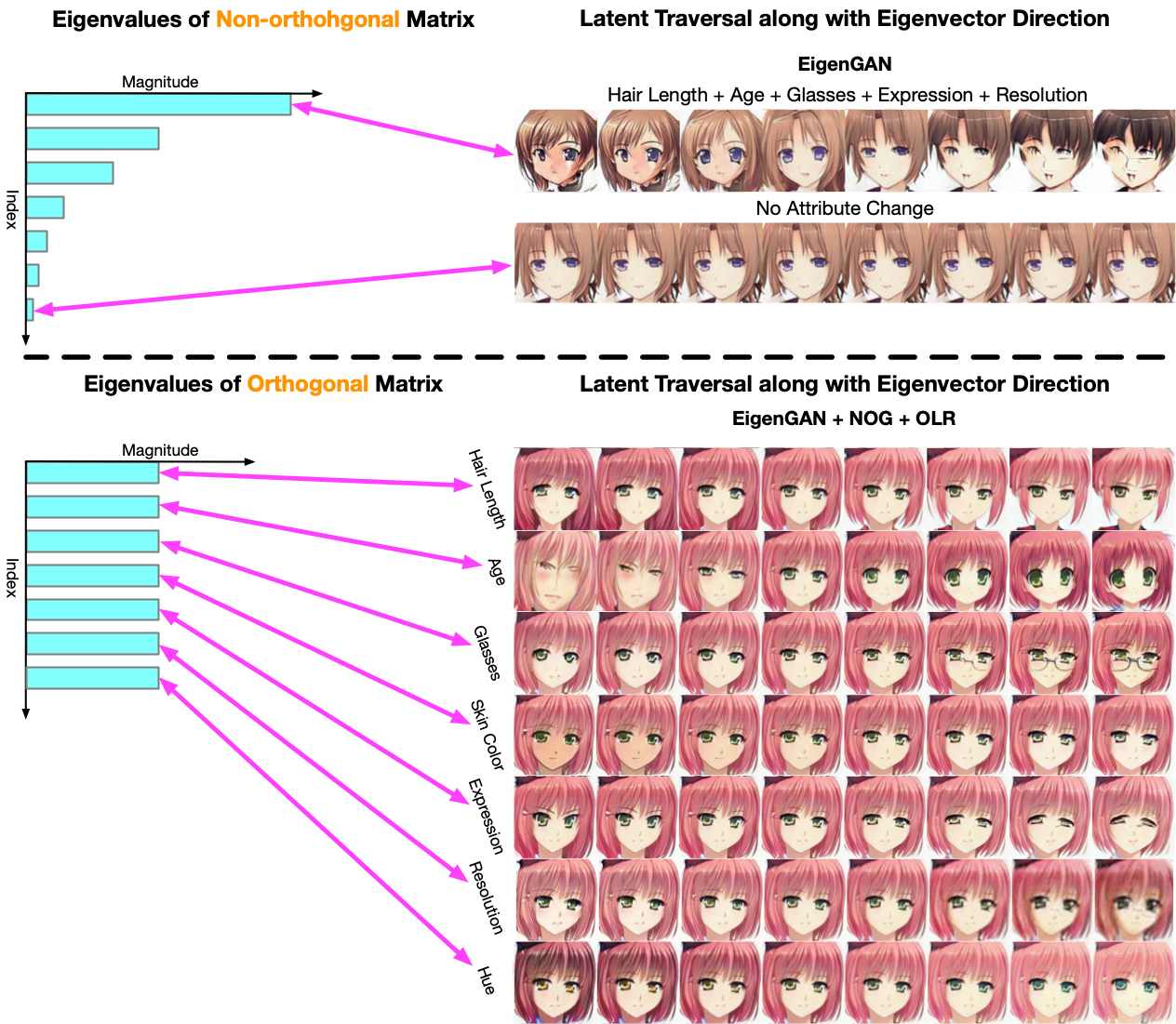}
    \caption{Illustration of the benefit of orthogonality in latent disentanglement. As revealed in~\cite{shen2021closed,zhu2021low}, the interpretable directions of latent codes are the eigenvectors of weight or gradient matrices. For non-orthogonal matrices, the principle eigenvector is of the most importance, which would make this direction correspond to many semantic attributes. The other eigenvectors might fail to capture any semantic information. By contrast, the eigenvectors of orthogonal matrices are equally important. The network with the orthogonal weight/gradient is likely to learn more disentangled representations. }
    \label{fig:ortho_illu}
\end{figure}

Though few previous works have applied implicit orthogonality as regularization in GANs~\cite{voynov2020unsupervised,peebles2020hessian,he2021eigengan,wei2021orthogonal}, there are no generally accepted explanations on how the orthogonality is related to the disentangled representations. 
Here we give an intuitive explanation. As discussed in the above image manipulation modelling, the eigenvectors of weight and gradient matrices naturally imply the interpretable directions for latent disentanglement. For common non-orthogonal matrices, the importance of each eigenvector is characterized by the corresponding eigenvalue. Each eigenvector is not equally important and the first few ones would dominate the spectrum. This imbalance would cause most semantic attributes entangled in the first few directions. Fig.~\ref{fig:ortho_illu} top illustrates this phenomenon: \emph{moving the latent code along with the top-1 eigenvector direction triggers changes of many semantic attributes. On the contrary, the small eigenvector direction does not indicate any semantic changes. The learned representation are thus deemed entangled.} 

The orthogonal matrices can greatly relieve this issue thanks to the flat spectrum and equally-important eigenvectors. As shown in Fig.~\ref{fig:ortho_illu} bottom, when our NOG and OLR are applied, each direction of the orthogonal matrix is equally important and corresponds to one semantic attribute. Shifting the latent code in one direction only changes the targeting semantic concept, while the identity and other attributes are not touched. Enforcing orthogonality would lead to the superior disentanglement of learned representations.


Our proposed NOG and OLR can serve as strict orthogonal gradient constraint and \textit{relaxed} orthogonal weight constraint, respectively. Enforcing them on the first layer after the latent code during the training process is very likely to lead to more disentangled representations. In Sec.~\ref{sec:latent}, we apply these two techniques in various GAN architectures and benchmarks for unsupervised latent disentanglement.

\section{Experiments}
\label{sec:exp}

\subsection{Covariance Conditioning}

We validate the proposed approaches in two applications: GCP and decorrelated BN. These two tasks are very representative because they have different usages of the SVD meta-layer. The GCP uses the matrix square root, while the decorrelated BN applies the inverse square root. In addition, the models of decorrelated BN often insert the SVD meta-layer at the beginning of the network, whereas the GCP models integrate the layer before the FC layer. 

\subsubsection{Decorrelated Batch Normalization}

\begin{figure}[htbp]
    \centering
    \includegraphics[width=0.4\linewidth]{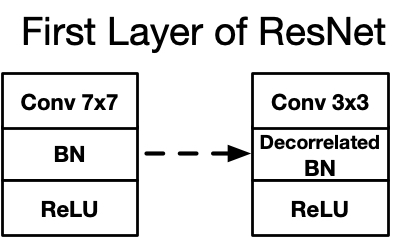}
    \caption{The scheme of the modified ResNet for decorrelated BN. We reduce the kernel size of the first convolution layer from $7{\times}7$ to $3{\times}3$. The BN after this layer is replaced with our decorrelated BN layer.}
    \label{fig:zca_arch}
\end{figure}

We use ResNet-50~\cite{he2016deep} as the backbone for the experiment on CIFAR10 and CIFAR100~\cite{krizhevsky2009learning}. The kernel size of the first convolution layer of ResNet is $7{\times}7$, which might not suit the low resolution of these two datasets (the images are only of size $32{\times}32$). To avoid this issue, we reduce the kernel size of the first convolution layer to $3{\times}3$. The stride is also decreased from $2$ to $1$. The BN layer after this layer is replace with our decorrelated BN layer (see Fig.~\ref{fig:zca_arch}). Let $\mX{\in}\mathbb{R}^{C{\times}BHW}$ denotes the reshaped feature. The whitening transform is performed as:
\begin{equation}
    \mX_{whitened} = (\mX\mX^{T})^{-\frac{1}{2}} \mX
\end{equation}
Compared with the vanilla BN that only standardizes the data, the decorrelated BN can further eliminate the data correlation between each dimension.

\begin{table}[ht]
    \centering
    \resizebox{0.99\linewidth}{!}{
    \begin{tabular}{r|c|c|c|c}
    \toprule
         \multirow{2}*{Methods} & \multicolumn{2}{c|}{CIFAR10} & \multicolumn{2}{c}{CIFAR100} \\
         \cline{2-5}
         &mean$\pm$std & min &mean$\pm$std & min\\
         \hline
         SVD &4.35$\pm$0.09&4.17&19.99$\pm$0.16&19.80 \\
         \hline
         SVD + SN &4.31$\pm$0.10 &4.15 &19.94$\pm$0.33 &19.60 \\
         SVD + OL &4.28$\pm$0.07 &4.23 &19.73$\pm$0.28 &19.54\\
         SVD + OW &4.42$\pm$0.09& 4.28&20.06$\pm$0.17&19.94\\
         \hline
         SVD +  NOG &\textbf{\textcolor{green}{4.15$\pm$0.06}}&\textbf{\textcolor{cyan}{4.04}} &\textbf{\textcolor{green}{19.43$\pm$0.24}} &\textbf{\textcolor{cyan}{19.15}} \\
         SVD + OLR  &\textbf{\textcolor{cyan}{4.23$\pm$0.17}}&\textbf{\textcolor{blue}{3.98}} &\textbf{\textcolor{cyan}{19.50$\pm$0.39}}&\textbf{\textcolor{green}{18.95}} \\
         \hline
         SVD + NOG + OW &\textbf{\textcolor{blue}{4.09$\pm$0.07}}& \textbf{\textcolor{green}{4.01}} &\textbf{\textcolor{blue}{19.22$\pm$0.28}}&\textbf{\textcolor{blue}{18.90}} \\
         SVD + NOG + OW + OLR &\textbf{\textcolor{red}{3.93$\pm$0.09}}&\textbf{\textcolor{red}{3.85}} &\textbf{\textcolor{red}{19.05$\pm$0.31}}&\textbf{\textcolor{red}{18.77}} \\
    \hline\hline
    NS iteration &4.20$\pm$0.11 &4.11  &19.45$\pm$0.33&19.01\\
    \bottomrule
    \end{tabular}
    }
    \caption{Performance comparison of decorrelated BN methods on CIFAR10/CIFAR100~\cite{krizhevsky2009learning} based on ResNet-50~\cite{he2016deep}. We report each result based on $10$ runs. The best four results are highlighted in \textbf{\textcolor{red}{red}}, \textbf{\textcolor{blue}{blue}}, \textbf{\textcolor{green}{green}}, and \textbf{\textcolor{cyan}{cyan}} respectively.}
    \label{tab:zca_res50}
\end{table}

Table~\ref{tab:zca_res50} compares the performance of each method on CIFAR10/CIFAR100~\cite{krizhevsky2009learning} based on ResNet-50~\cite{he2016deep}. Both of our NOG and OLR achieve better performance than other weight treatments and the SVD. Moreover, when hybrid treatments are adopted, we can observe step-wise steady improvements on the validation errors. Among these techniques, the joint usage of OLR with NOG and OW achieves the best performances across metrics and datasets, outperforming the SVD baseline by $0.4\%$ on CIFAR10 and by $0.9\%$ on CIFAR100. This demonstrates that these treatments are complementary and can benefit each other.   

\begin{table}[ht]
    \centering
    \resizebox{0.99\linewidth}{!}{
    \begin{tabular}{r|c|c|c|c}
    \toprule
         \multirow{2}*{Methods} & \multicolumn{2}{c|}{DenseNet-121~\cite{huang2017densely}} & \multicolumn{2}{c}{MobileNet-v2~\cite{howard2017mobilenets}} \\
         \cline{2-5}
         &mean$\pm$std & min &mean$\pm$std & min\\
         \hline
         SVD &27.37$\pm$0.54&26.88 & 34.35$\pm$0.32 &34.00 \\
         \hline
          SVD + SN &27.05$\pm$0.44 &26.51 & 34.19$\pm$0.37 & 33.82 \\
         SVD + OL &27.41$\pm$0.35 &26.99 &34.58$\pm$0.43 &34.15  \\
         SVD + OW & 27.25$\pm$0.47 &26.67 &34.27$\pm$0.46 & 33.77\\
         \hline
         SVD + NOG & \textbf{\textcolor{green}{25.14$\pm$0.39}} & \textbf{\textcolor{green}{24.65}} & \textbf{\textcolor{green}{33.42$\pm$0.41}} & \textbf{\textcolor{cyan}{32.91}} \\
         SVD + OLR  & \textbf{\textcolor{cyan}{25.34$\pm$0.28}} & \textbf{\textcolor{cyan}{25.01}} &\textbf{\textcolor{cyan}{33.59$\pm$0.64}} &\textbf{\textcolor{green}{32.84}} \\
         \hline
         SVD + NOG + OW & \textbf{\textcolor{blue}{24.49$\pm$0.43}} &\textbf{\textcolor{blue}{23.97}} &\textbf{\textcolor{blue}{33.13$\pm$0.55}} & \textbf{\textcolor{blue}{32.61}} \\
         SVD + NOG + OW + OLR &\textbf{\textcolor{red}{23.74$\pm$0.24}} &\textbf{\textcolor{red}{23.41}} &\textbf{\textcolor{red}{32.83$\pm$0.48}} & \textbf{\textcolor{red}{32.33}} \\
    \hline\hline
    NS iteration &25.87$\pm$0.43 &25.31 &33.67$\pm$0.51 &33.24\\
    \bottomrule
    \end{tabular}
    }
    \caption{Performance comparison of decorrelated BN methods on CIFAR100~\cite{krizhevsky2009learning} with DenseNet-121~\cite{huang2017densely} and MobileNet-v2~\cite{howard2017mobilenets} based on $10$ runs. The best four results are highlighted in \textbf{\textcolor{red}{red}}, \textbf{\textcolor{blue}{blue}}, \textbf{\textcolor{green}{green}}, and \textbf{\textcolor{cyan}{cyan}} respectively.}
    \label{tab:zca_dense_mobile}
\end{table}

\begin{figure}[htbp]
    \centering
    \includegraphics[width=0.99\linewidth]{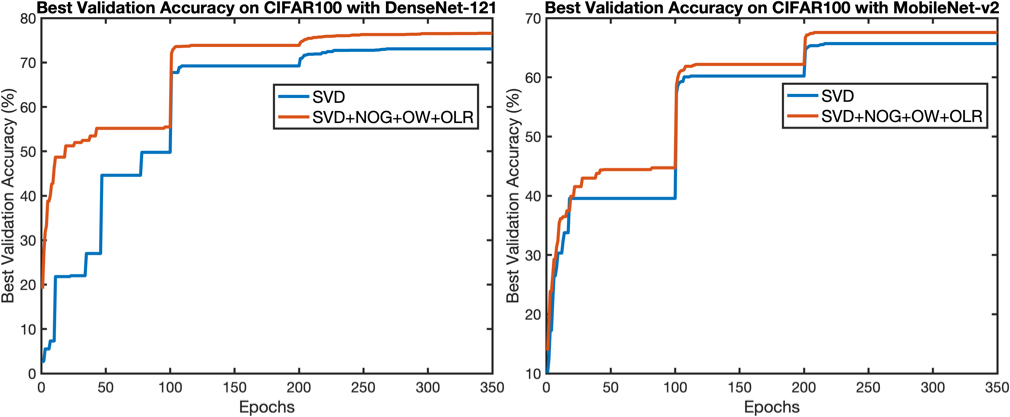}
    \caption{{The best validation accuracy during the training process. Our proposed techniques can consistently improve the convergence speed and help the model to achieve better accuracy within fewer training epochs.}}
    \label{fig:cvgc_improvement}
\end{figure}

{Table~\ref{tab:zca_dense_mobile} presents the validation errors on CIFAR100 with DenseNet-121~\cite{huang2017densely} and MobileNet-v2~\cite{howard2017mobilenets}. The results are coherent with those on ResNet-50~\cite{he2016deep}: our methods bring consistent performance improvements to the ordinary SVD on different architectures. This demonstrates the model-agnostic property of the proposed orthogonality approaches. Fig.~\ref{fig:cvgc_improvement} displays the corresponding best validation accuracy during the training process. Our method can also accelerate the convergence of the training process. The acceleration is particularly significant in the initial training stage. }

{Finally, we would like to note that the performance gain of our methods depends on the specific architectures and the ill-conditioned extent of the covariance. Generally speaking, the larger the model is, the worse-conditioned the covariance is and the larger the performance gain would be. Take the above decorrelated BN experiments as an example, the accuracy improvement on MobileNet is around $1.5\%$, while the performance gain on larger DenseNet is about $4.0\%$.}

\subsubsection{Global Covariance Pooling}

\begin{figure}[htbp]
    \centering
    \includegraphics[width=0.7\linewidth]{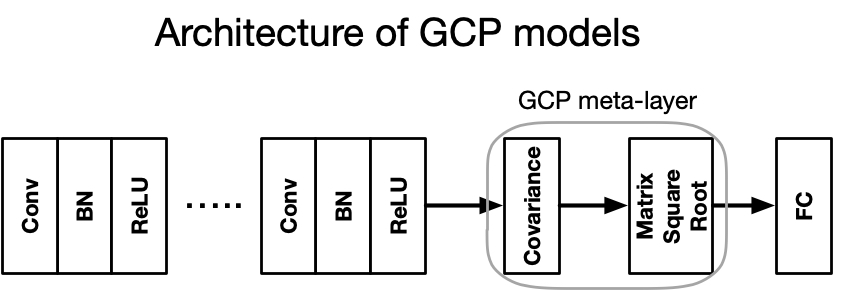}
    \caption{The architecture of a GCP model~\cite{li2017second,song2021approximate}. After all the convolution layers, the covariance square root of the feature is computed and used as the final representation.}
    \label{fig:gcp_arch}
\end{figure}

We use ResNet-18~\cite{he2016deep} for the GCP experiment and train it from scratch on ImageNet~\cite{deng2009imagenet}. Fig.~\ref{fig:gcp_arch} displays the overview of a GCP model. For the ResNet backbone, the last Global Average Pooling (GAP) layer is replaced with our GCP layer. Consider the final batched convolutional feature $\mX{\in}\mathbb{R}^{B{\times}C{\times}HW}$. We compute the matrix square root of its covariance as:
\begin{equation}
    \mQ = (\mX\mX^{T})^{\frac{1}{2}}
\end{equation}
where $\mQ{\in}\mathbb{R}^{B{\times}C{\times}C}$ is used as the final representation and directly passed to the fully-connected (FC) layer.   

\begin{table}[htbp]
    \centering
    \caption{Performance comparison of different GCP methods on ImageNet~\cite{deng2009imagenet} based on ResNet-18~\cite{he2016deep}. The failure times denote the total times of non-convergence of the SVD solver during one training process. The best four results are highlighted in \textbf{\textcolor{red}{red}}, \textbf{\textcolor{blue}{blue}}, \textbf{\textcolor{green}{green}}, and \textbf{\textcolor{cyan}{cyan}} respectively.}
    \resizebox{0.99\linewidth}{!}{
    \begin{tabular}{r|c|c|c}
    \toprule
        Method & \makecell[c]{Failure\\ Times} & Top-1 Acc. (\%) & Top-5 Acc. (\%) \\
    \hline
        SVD & 5 & 73.13 & 91.02 \\
    \hline
        SVD + SN &2 &73.28 ($\uparrow$ 0.2) &91.11 ($\uparrow$ 0.1)\\
        SVD + OL & 1& 71.75 ($\downarrow$ 1.4) &90.20 ($\downarrow$ 0.8)\\
        SVD + OW &2 &73.07 ($\downarrow$ 0.1) &90.93 ($\downarrow$ 0.1)\\
    \hline
        SVD + NOG & 1 &\textbf{\textcolor{green}{73.51}} (\textbf{\textcolor{green}{$\uparrow$ 0.4}}) & \textbf{\textcolor{green}{91.35}} (\textbf{\textcolor{green}{$\uparrow$ 0.3}})\\
        SVD + OLR & 0 & \textbf{\textcolor{cyan}{73.39}} (\textbf{\textcolor{cyan}{$\uparrow$ 0.3}}) & \textbf{\textcolor{cyan}{91.26}} (\textbf{\textcolor{cyan}{$\uparrow$ 0.2}})\\
    \hline
        SVD + NOG + OW & 0  & \textbf{\textcolor{blue}{73.71}} (\textbf{\textcolor{blue}{$\uparrow$ 0.6}})& \textbf{\textcolor{blue}{91.43}} (\textbf{\textcolor{blue}{$\uparrow$ 0.4}})\\
        SVD + NOG + OW + OLR & 0  & \textbf{\textcolor{red}{73.82}} (\textbf{\textcolor{red}{$\uparrow$ 0.7}})& \textbf{\textcolor{red}{91.57}} (\textbf{\textcolor{red}{$\uparrow$ 0.6}})\\
    \hline\hline
    NS iteration & 0 & 73.36 ($\uparrow$ 0.2) & 90.96 ($\downarrow$ 0.1)\\
    \bottomrule
    \end{tabular}
    }
    \label{tab:gcp_res18}
\end{table}

Table~\ref{tab:gcp_res18} presents the total failure times of the SVD solver in one training process and the validation accuracy on ImageNet~\cite{deng2009imagenet} based on ResNet-18~\cite{he2016deep}. The results are very coherent with our experiment of decorrelated BN. Among the weight treatments, the OL and OW hurt the performance, while the SN improves that of SVD by $0.2\%$. Our proposed NOG and OLR outperform the weight treatments and improve the SVD baseline by $0.4\%$ and by $0.3\%$, respectively. Moreover, the combinations with the orthogonal weight further boost the performance. Specifically, combining NOG and OW surpasses the SVD by $0.6\%$. The joint use of OW with NOG and OLR achieves the best performance among all the methods and beats the SVD by $0.7\%$.

\begin{figure*}[t]
    \centering
    \includegraphics[width=0.99\linewidth]{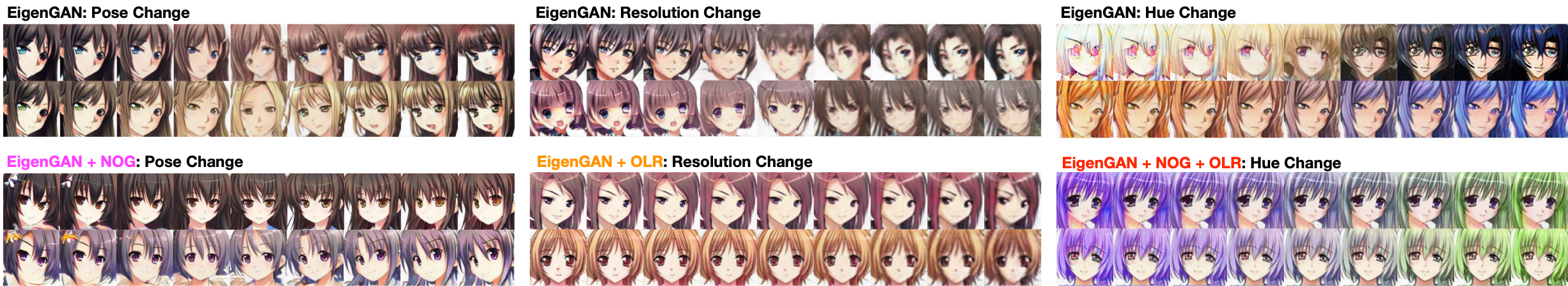}
    \caption{Latent traversal on AnimeFace~\cite{chao2019/online}. The EigenGAN has entangled attributes in the identified interpretable directions, while our methods achieve better disentanglement and each direction corresponds to a unique attribute.  }
    \label{fig:animeface_eigengan_compare}
\end{figure*}

\begin{figure}[htbp]
    \centering
    \includegraphics[width=0.99\linewidth]{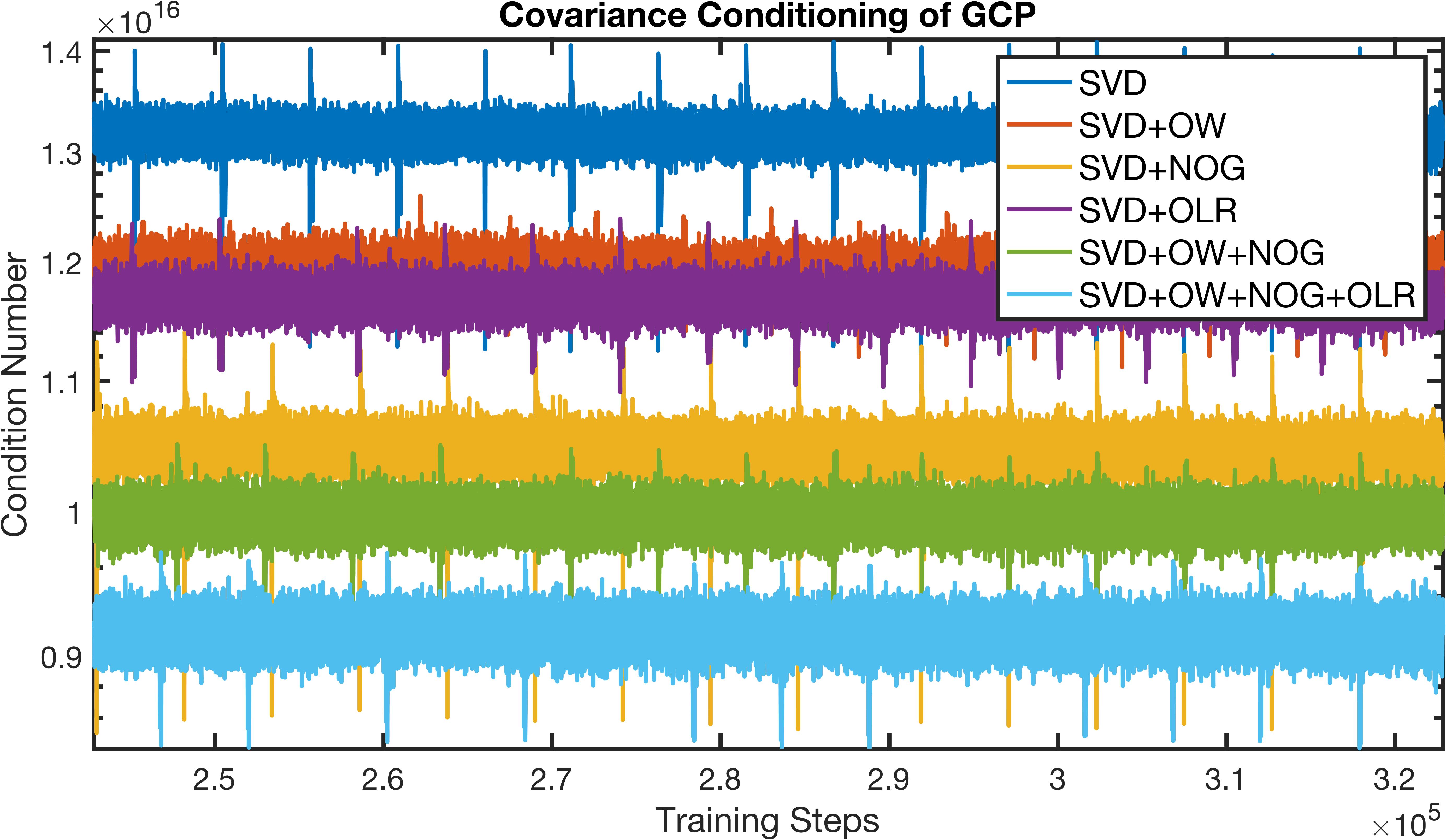}
    \caption{The covariance conditioning of GCP methods in the later stage of the training. The periodic spikes are caused by the evaluation on the validation set after every epoch.}
    \label{fig:gcp_cond}
\end{figure}

Fig.~\ref{fig:gcp_cond} depicts the covariance conditioning in the later training stage. Our OLR and the OW both reduce the condition number by around $1e15$, whereas the proposed NOG improves the condition number by $2e15$. When hybrid treatments are used, combining NOG and OW attains better conditioning than the separate usages. Furthermore, simultaneously using all the techniques leads to the best conditioning and improves the condition number by $5e15$. 

The covariance conditioning of GCP tasks is not improved as much as that of decorrelated BN. This might stem from the unique architecture of GCP models: the covariance is directly used as the final representation and fed to the FC layer. We conjecture that this setup might cause the covariance to have a high condition number. The approximate solver (NS iteration) does not have well-conditioned matrices either (${\approx}1e15$), which partly supports our conjecture.

\subsubsection{Computational Cost}

\begin{table}[htbp]
    \centering
    \caption{Time consumption of each forward pass (FP) and backward pass (BP) measured on a RTX A6000 GPU. The evaluation is based on ResNet-50 and CIFAR100.} 
    \resizebox{0.8\linewidth}{!}{
    \begin{tabular}{c|cc}
    \toprule
         Methods & FP (ms) & BP (ms) \\
    \midrule
         SVD & 44 & 95 \\
         SVD + NOG & 44 & 97 (+2)\\
         SVD + OLR & 44 & 96 (+1) \\
         SVD + OW & 48 (+4) & 102 (+7) \\
         SVD + OW + NOG + OLR & 49 (+5) &106 (+11) \\
         NS Iteration & 43 & 93\\
         \midrule
         Vanilla ResNet-50 & 42  & 90 \\
    \bottomrule
    \end{tabular}
    }
    \label{tab:time}
\end{table}

Table~\ref{tab:time} compares the time consumption of a single training step for the experiment of decorrelated BN. Our NOG and OLR bring negligible computational costs to the BP ($2\%$ and $1\%$), while the FP is not influenced. Even when all techniques are applied, the overall time costs are marginally increased by $10\%$. Notice that NOG and OLR have no impact on the inference speed.

\subsection{Latent Disentanglement}
\label{sec:latent}

In this subsection, we first introduce the experiment setup, followed by the evaluation results on different GAN architectures and datasets. We defer the implementation details to the Supplementary Material.

\subsubsection{Experimental Setup}

\noindent\textbf{Models.} We evaluate our methods on EigenGAN~\cite{he2021eigengan} and vanilla GAN~\cite{goodfellow2014generative}. EigenGAN~\cite{he2021eigengan} is a particular GAN architecture dedicated to latent disentanglement. It progressively injects orthogonal subspaces into each layer of the generator, which can mine controllable semantic attributes in an unsupervised manner. For the vanilla GAN~\cite{goodfellow2014generative}, we adopt the basic GAN model that consists of stacked convolutional layers and do not make any architectural modifications.

\noindent\textbf{Datasets.} For EigenGAN, we use AnimeFace~\cite{chao2019/online} and FFHQ~\cite{kazemi2014one} datasets. AnimeFace~\cite{chao2019/online} is comprised of $63,632$ aligned anime faces with resolution varying from $90{\times}90$ to $120{\times}120$. FFHQ~\cite{kazemi2014one} consists of $70,000$ high-quality face images that have considerable variations in identifies and have good coverage in common accessories. Since the vanilla GAN has a smaller architecture and fewer parameters, we use relatively simpler CelebA~\cite{liu2018large} and LSUN Church~\cite{yu2015lsun} datasets. CelebA~\cite{liu2018large} contains $202,599$ face images of $10,177$ celebrities, while LSUN Church~\cite{yu2015lsun} has $126,227$ scenes images of church.


\noindent\textbf{Metrics.} We use Frechet Inception Distance (FID)~\cite{heusel2017gans} to quantitatively evaluate the quality of generate images. For the performance of latent disentanglement, we use Variational Predictability (VP)~\cite{zhu2020learning} as the quantitative metric. The VP metric adopts the few-shot learning setting to measure the generalization abilities of a simple neural network in classifying the discovered latent directions.

\noindent\textbf{Baselines.} For the EigenGAN model that already has inherent orthogonality constraints and good disentanglement abilities, we compare the ordinary EignGAN with the modified version augmented by our proposed orthogonal techniques (NOG and OLR). For the vanilla GAN that suffers from limited disentanglement, we compare our NOG and OLR against other disentanglement schemes used in GANs, including (1) Hessian Penalty (HP)~\cite{peebles2020hessian}, (2) Orthogonal Jacobian Regularization (OrthoJar)~\cite{wei2021orthogonal}, and (3) Latent Variational Predictability (LVP)~\cite{zhu2020learning}.

\subsubsection{EigenGAN Architecture and Modifications}


\begin{figure*}[t]
    \centering
    \includegraphics[width=0.99\linewidth]{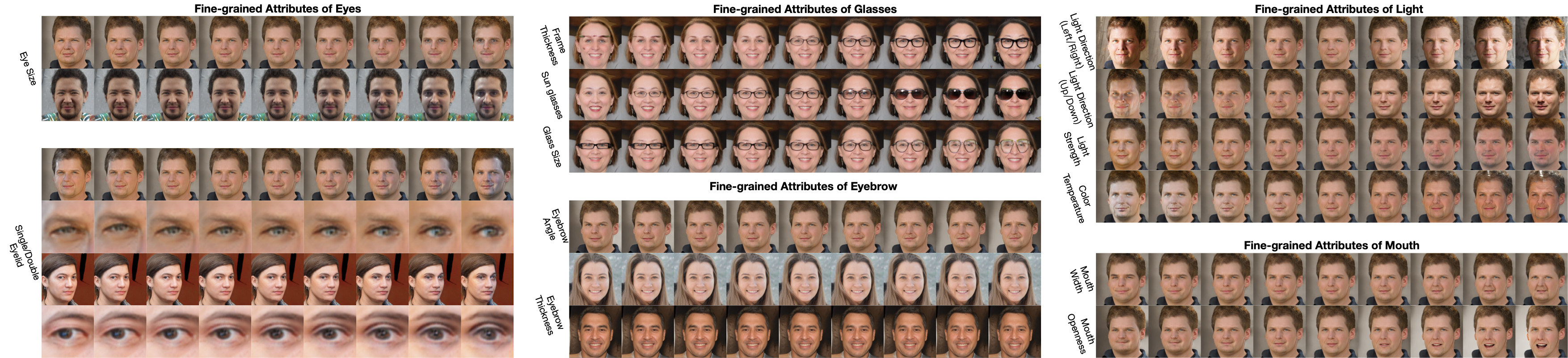}
    \caption{Visualization of some fine-grained attributes learned by out method on FFHQ~\cite{kazemi2014one} dataset. Our method can learn very subtle and fine-grained attributes while keeping the identity unchanged.}
    \label{fig:ffhq_finegrained}
\end{figure*}

\begin{figure}[h]
    \centering
    \includegraphics[width=0.9\linewidth]{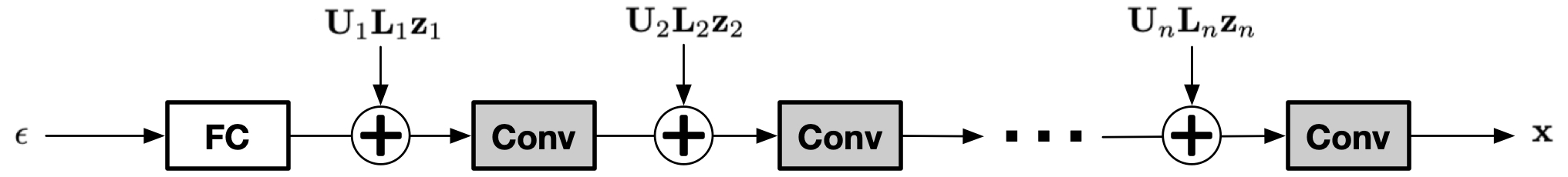}
    \caption{Overview of the EigenGAN architecture. }
    \label{fig:eigengan_arch}
\end{figure}

Fig.~\ref{fig:eigengan_arch} displays the overview of the EigenGAN. At each layer, the latent code $\mathbf{z}_{i}$ is multiplied with the orthogonal basis $\mathbf{U}_{i}$ and the diagonal importance matrix $\mathbf{L}_{i}$ to inject weighted orthogonal subspace for disentangled representation learning. The original EigenGAN~\cite{he2021eigengan} adopts the OL loss $||\mathbf{U}_{i}\mathbf{U}_{i}^{T}{-}\mathbf{I}||_{\rm F}$ to enforce \emph{relaxed} orthogonality to each subspace $\mathbf{U}_{i}$. Instead, we apply our NOG and OLR to achieve the weight and gradient orthogonality, respectively. {Notice that when our NOG and OLR are applied, we do not use the OL loss of EigenGAN. This is because the \emph{soft} orthogonality introduced by the OL loss might not be enforced under the gradient manipulation of our NOG, which is similar to our experimental results of decorrelated BN (see Sec.~\ref{sec:ol_nog_combination}).}

\subsubsection{Results on EigenGAN}

\begin{figure}[htbp]
    \centering
    \includegraphics[width=0.99\linewidth]{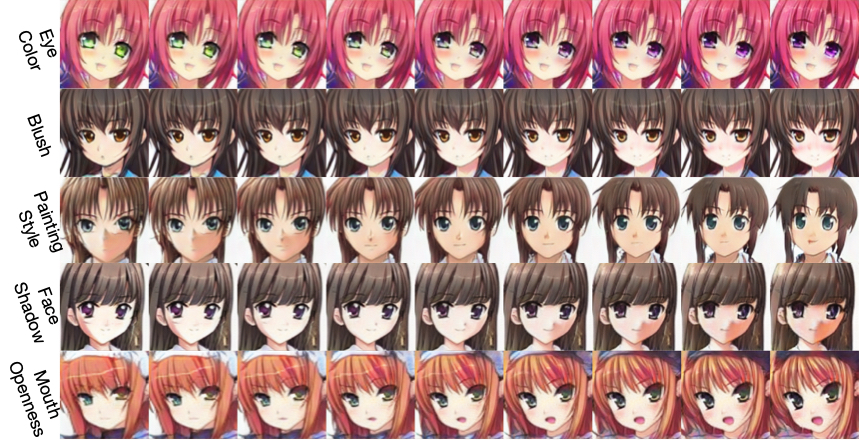}
    \caption{Subtle semantic attributes mined by our method. }
    \label{fig:animeface_eigengan_finegrained}
\end{figure}

\noindent\textbf{Qualitative Evaluation.} Fig.~\ref{fig:animeface_eigengan_compare} compares the latent traversal results of the ordinary EigenGAN and our methods on AnimeFace. The interpretable direction of EigenGAN has many entangled attributes; the identity is poorly preserved during the latent traversal. By contrast, moving along with the discovered direction of our method would only introduce changes of a single semantic attribute. This demonstrates that our interpretable directions have more precisely-controlled semantics and our orthogonality techniques indeed help the model to learn more disentangled representations. Moreover, thanks to the power of orthogonality, our methods can mine many subtle and fine-grained attributes. Fig.~\ref{fig:animeface_eigengan_finegrained} displays such attributes that are precisely captured by out method but are not learnt by EigenGAN. These attributes include very subtle local details of the image, such as facial blush, facial shadow, and mouth openness. 

\begin{figure}[tbp]
    \centering
    \includegraphics[width=0.99\linewidth]{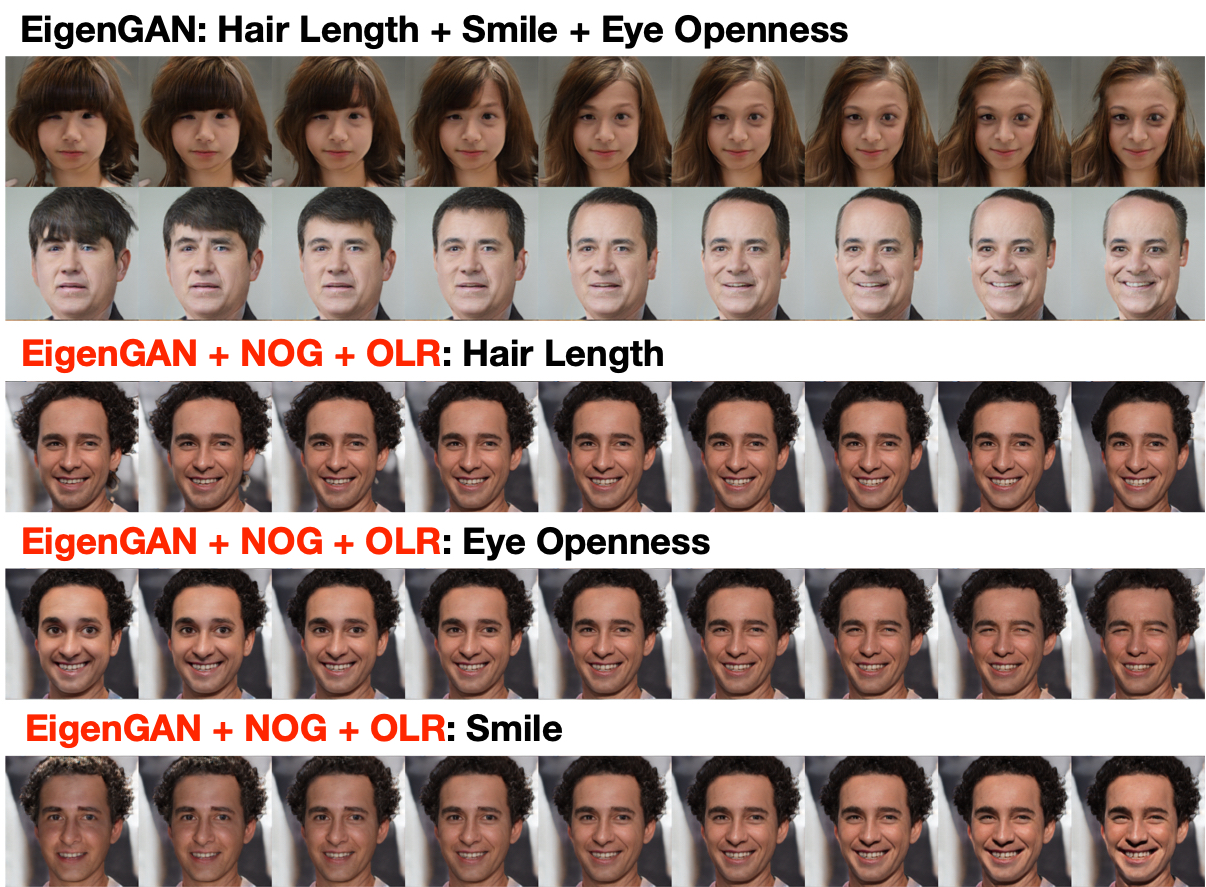}
    \caption{Qualitative comparison on FFHQ. The attributes are entangled in one latent direction of EigenGAN, while our method can avoid this and discover orthogonal concepts.}
    \label{fig:ffhq_quantiative}
\end{figure}

Fig.~\ref{fig:ffhq_quantiative} compares the exemplary latent traversal on FFHQ. Similar with the result on AnimeFace, the interpretable directions have more disentangled attributes when our orthogonality techniques are used. Since FFHQ covers a wide range of image attributes, our method is able to learn very fine-grained attributes (\emph{e.g.,} angle and thickness of eyebrow) of a given super attribute (\emph{e.g.,} eyebrow) accordingly. We give a few examples in Fig.~\ref{fig:ffhq_finegrained}. As can be observed, our method can precisely control the subtle detail of the image while keeping other attributes unchanged.

\begin{table}[htbp]
    \centering
    \resizebox{0.99\linewidth}{!}{
    \begin{tabular}{c|cc|cc}
    \toprule
        \multirow{2}*{Methods} & \multicolumn{2}{c|}{AnimeFace~\cite{chao2019/online}} & \multicolumn{2}{c}{FFHQ~\cite{kazemi2014one}} \\
        \cmidrule{2-5}
         & FID ($\downarrow$) & VP ($\uparrow$) & FID ($\downarrow$) & VP ($\uparrow$)\\
    \midrule
         EigenGAN & 23.59 & 37.01 & 36.81 & 31.79\\
    \midrule
         EigenGAN+NOG &19.48 &43.53 &33.34 &37.27 \\
         EigenGAN+OLR &18.30 &43.99 &31.42 &37.23 \\
         EigenGAN+OLR+NOG &\textbf{16.31} &\textbf{45.48} &\textbf{30.06} &\textbf{39.32} \\
    \bottomrule
    \end{tabular}
    }
    \caption{Quantitative evaluation on EigenGAN.}
    \label{tab:eigengan_results}
\end{table}

\noindent\textbf{Quantitative Evaluation.} Table~\ref{tab:eigengan_results} compares the performance of EigenGAN on AnimeFace and FFHQ datasets. Our proposed NOG and OLR can improve both the image quality score (FID) and the disentanglement score (VP). Furthermore, when these two techniques are combined, the evaluation results achieve the best performance across metrics and datasets. This implies that enforcing simultaneous gradient and weight orthogonality allows for the learning of more disentangled representations and improved image fidelity. 



\noindent\textbf{Discussion.} Both quantitative and qualitative evaluation on two datasets demonstrates that our orthogonality approaches lead to better latent disentanglement than the inherent orthogonality loss of EigenGAN. This behavior is coherent to our previous experiment of decorrelated BN: the proposed NOG and OLR also outperform OL in that case. This further confirms the general applicability of our orthogonal methods.


\subsubsection{Vanilla GAN Architecture} 
For the vanilla GAN model, we use simple convolutional layers as building blocks. The orthogonality techniques are applied on the first convolution layer after the latent code.   

\subsubsection{Results on Vanilla GAN}

\begin{figure}[htbp]
    \centering
    \includegraphics[width=0.99\linewidth]{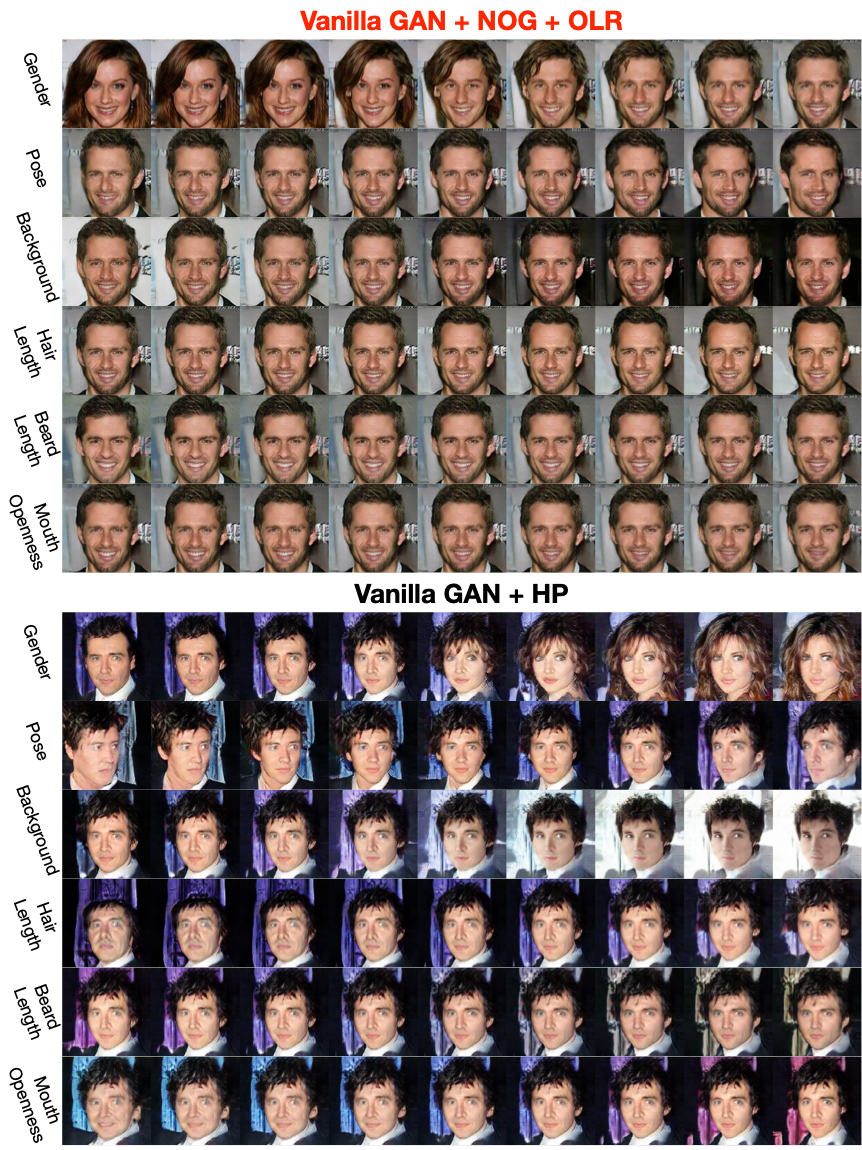}
    \caption{Qualitative comparison on CelebA. For HP~\cite{peebles2020hessian}, The latent traversal in one direction would introduce many attributes changes. By contrast, the image identity of our method is well preserved and only the target attribute varies.}
    \label{fig:celeba}
\end{figure}

\begin{figure}[htbp]
    \centering
    \includegraphics[width=0.99\linewidth]{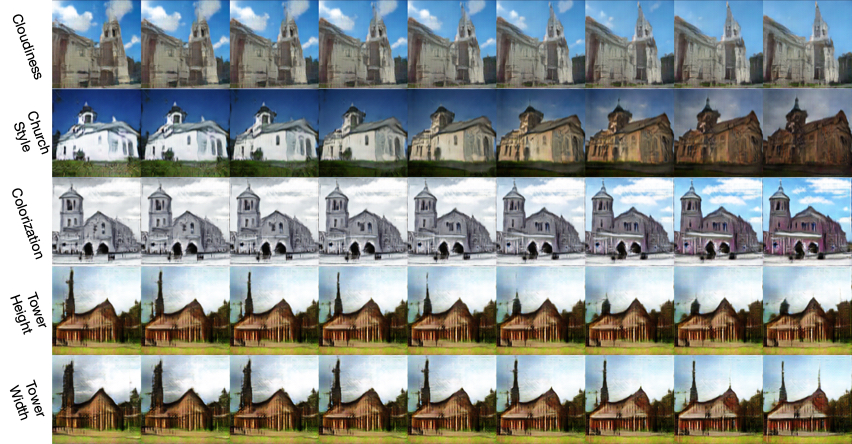}
    \caption{Latent traversal of our NOG on LSUN Church.}
    \label{fig:lsun}
\end{figure}

\noindent\textbf{Qualitative Evaluation.} Fig.~\ref{fig:celeba} presents the qualitative evaluation results on CelebA~\cite{liu2018large} against HP~\cite{peebles2020hessian}. The semantic factors discovered by our methods controls traversal process more precisely; only a single attribute is changed when one latent code is modified. By contrast, a interpretable direction mined by HP~\cite{peebles2020hessian} would correspond to multiple attributes sometimes. This implies that the learned representations and attributes of our NOG and OLR are more disentangled. Fig.~\ref{fig:lsun} displays some learned attributes of our methods. The complex scenes and structures of churches are preserved well, and each semantic factor precisely controls the image attribute. This also demonstrates the diverse application domains of our disentanglement method beyond face analysis. 

\begin{table}[htbp]
    \centering
    \resizebox{0.99\linewidth}{!}{
    \begin{tabular}{c|c|cc|cc}
    \toprule
        \multirow{2}*{Methods} &\multirow{2}*{Time (ms)} & \multicolumn{2}{c|}{CelebA~\cite{liu2018large}} & \multicolumn{2}{c}{LSUN Church~\cite{yu2015lsun}} \\
        \cmidrule{3-6}
        & & FID ($\downarrow$) & VP ($\uparrow$) & FID ($\downarrow$) & VP ($\uparrow$)\\
    \midrule
        OrJar~\cite{wei2021orthogonal} &23 & 32.43 & 24.24 & 38.96 & 11.62   \\
        HP~\cite{peebles2020hessian}    &30 & 31.65 & 24.67 & 39.20 & \textbf{\textcolor{green}{13.73}} \\
        LVP~\cite{zhu2020learning}    &16 & 34.36 & 23.49 & 41.24 & 12.58\\
    \midrule
        NOG   &\textbf{8} &\textbf{\textcolor{red}{29.69}} &\textbf{\textcolor{green}{25.33}} & \textbf{\textcolor{green}{37.22}} & 13.43 \\
        OLR   &\textbf{8} &\textbf{\textcolor{green}{33.29}}&\textbf{\textcolor{blue}{27.22}} & \textbf{\textcolor{blue}{37.83}} & \textbf{\textcolor{blue}{14.50}} \\
        NOG+OLR & \textbf{9} &\textbf{\textcolor{blue}{30.65}} &\textbf{\textcolor{red}{28.74}} & \textbf{\textcolor{red}{35.20}} & \textbf{\textcolor{red}{16.98}} \\
    \bottomrule
    \end{tabular}
    }
    \caption{Quantitative evaluation on vanilla GAN. We measure the time consumption of a single forward pass and backward pass. The best three results are highlighted in \textbf{\textcolor{red}{red}}, \textbf{\textcolor{blue}{blue}}, and \textbf{\textcolor{green}{green}} respectively.}
    \label{tab:vanillagan_results}
\end{table}

\noindent\textbf{Quantitative Evaluation.} Table.~\ref{tab:vanillagan_results} reports the quantitative evaluation results on vanilla GAN. Our proposed orthogonality techniques outperform other disentanglement schemes in terms of both FID and VP, achieving state-of-the-art performance in the unsupervised latent disentanglement. Moreover, our approaches are much more efficient than other baselines due to the marginal computational cost.




\begin{table}[htbp]
    \centering
    \resizebox{0.99\linewidth}{!}{
    \begin{tabular}{c|ccc|ccc}
    \toprule
         Datasets & OrJar~\cite{wei2021orthogonal} & HP~\cite{peebles2020hessian} & LVP~\cite{zhu2020learning} & NOG & OLR & NOG+OLR\\
         \midrule
         CelebA~\cite{liu2018large}&2.75 &2.67 &2.78 &2.28 &2.14 &\textbf{2.01} \\
         Church~\cite{yu2015lsun}&2.48 &2.57 &2.66 &2.13 &2.09 &\textbf{1.93} \\
    \bottomrule
    \end{tabular}
    }
    \caption{{Condition number of the first convolution weight in vanilla GANs on CelebA~\cite{liu2018large} and LSUN Church~\cite{yu2015lsun}.}}
    \label{tab:cond_gan}
\end{table}

\noindent\textbf{{Condition Number in Vanilla GANs.}} {Similar to our previous experiments, we measure the condition number of the fist convolution weight in vanilla GANs (\emph{i.e.,} the projection matrix that maps latent codes to features). Table~\ref{tab:cond_gan} presents the evaluation results on CelebA~\cite{liu2018large} and LSUN Church~\cite{yu2015lsun}. As can be observed, our methods (NOG, OLR, and NOG+OLR) outperform other baselines and have much better condition numbers. This demonstrates that our methods can also improve the conditioning of the weight matrix of vanilla GANs. Notice that the convolution weight matrix is small in dimensionality. The corresponding condition number is thus much smaller compared with the covariance conditioning in the previous experiments.}

\section{Conclusion}
\label{sec:conclusion}

In this paper, we explore different approaches to improve the covariance conditioning of the SVD meta-layer. Existing treatments on orthogonal weight are first studied. Our experiments reveal that these techniques could improve the conditioning but might hurt the performance due to the limitation on the representation power. To avoid the side effect of orthogonal weight, we propose the nearest orthogonal gradient and the optimal learning rate, both of which could simultaneously attain better covariance conditioning and improved generalization abilities. Moreover, their combinations with orthogonal weight further boost the performance. Besides the usage on the SVD meta-layer, we show that our proposed orthogonality approaches can benefit generative models for better latent disentanglement.

\section*{Acknowledgements}
This research was supported by the EU H2020 projects AI4Media (No. 951911) and SPRING (No. 871245) and by the PRIN project CREATIVE Prot. 2020ZSL9F9.

%


\ifCLASSOPTIONcaptionsoff
  \newpage
\fi



%
\bibliographystyle{IEEEtran}
\bibliography{egbib}
%

%

\begin{IEEEbiography}[{\includegraphics[width=1in,height=1.25in,keepaspectratio]{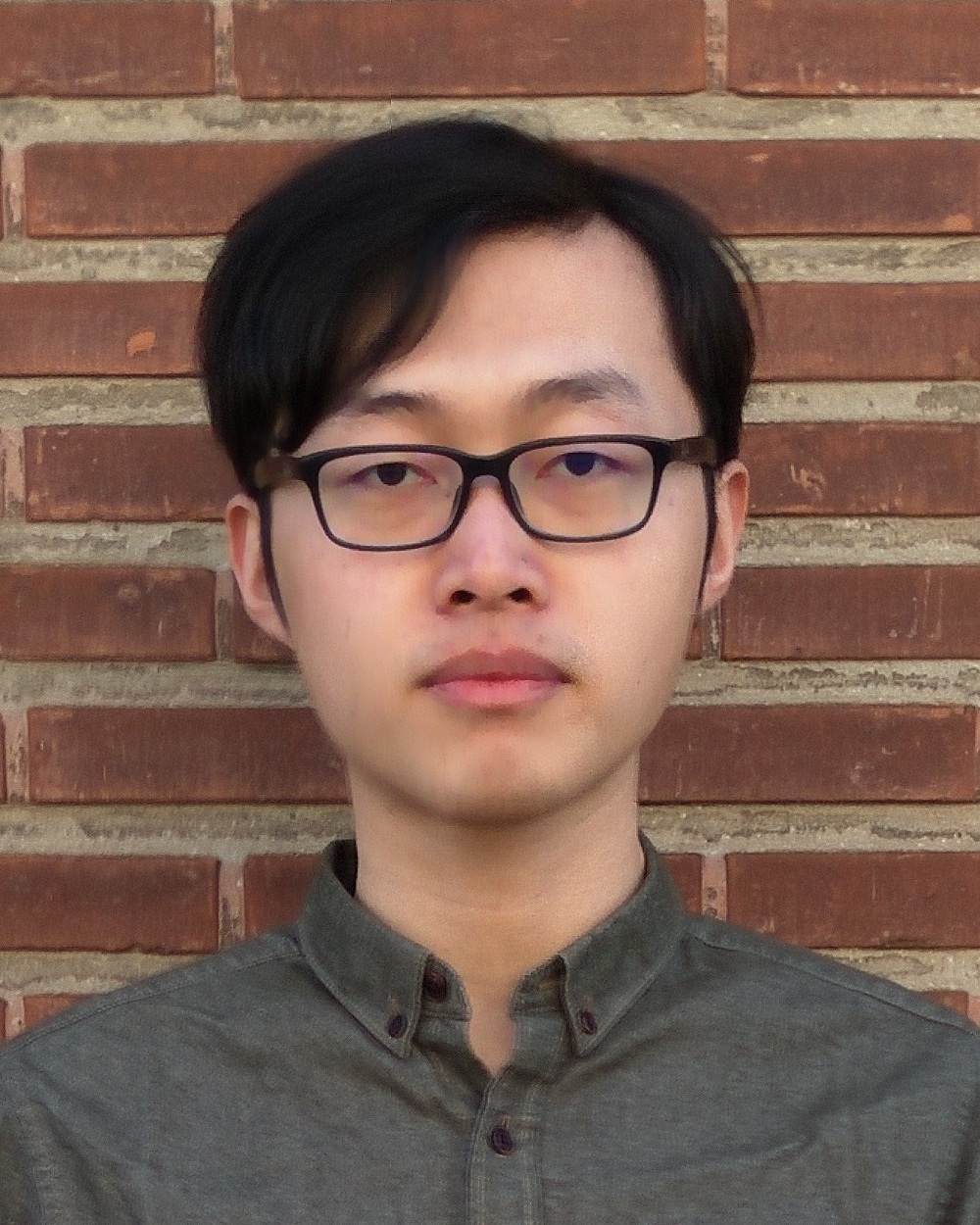}}]{Yue Song}
received the B.Sc. \emph{cum laude} from KU Leuven, Belgium and the joint M.Sc. \emph{summa cum laude} from the University of Trento, Italy and KTH Royal Institute of Technology, Sweden. Currently, he is a Ph.D. student with the Multimedia and Human Understanding Group (MHUG) at the University of Trento, Italy. His research interests are computer vision, deep learning, and numerical analysis and optimization.
\end{IEEEbiography}

\begin{IEEEbiography}[{\includegraphics[width=1in,height=1.25in,keepaspectratio]{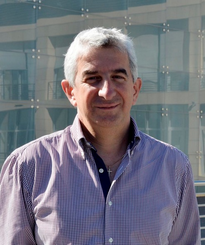}}]{Nicu Sebe} is Professor with the University of
Trento, Italy, leading the research in the areas
of multimedia information retrieval and human
behavior understanding. He was the General
Co- Chair of ACM Multimedia 2013, and the
Program Chair of ACM Multimedia 2007 and
2011, ECCV 2016, ICCV 2017 and ICPR 2020.
He is a fellow of the International Association for
Pattern Recognition.
\end{IEEEbiography}


\begin{IEEEbiography}[{\includegraphics[width=1in,height=1.25in,keepaspectratio]{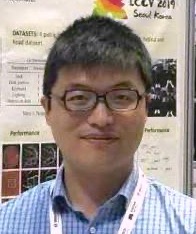}}]{Wei Wang}
is an Assistant Professor of Computer Science at University of Trento, Italy. Previously, after obtaining his PhD from University of
Trento in 2018, he became a Postdoc at EPFL,
Switzerland. His research interests include machine learning and its application to computer
vision and multimedia analysis.
\end{IEEEbiography}

\appendices
\newpage

\section{Background: SVD Meta-Layer}

This subsection presents the background knowledge about the propagation rules of the SVD meta-layer.

\subsection{Forward Pass} 

Given the reshape feature $\mX{\in}\mathbb{R}^{d{\times}N}$ where $d$ denotes the feature dimensionality (\emph{i.e.,} the number of channels) and $N$ represents the number of features (\emph{i.e.,} the product of spatial dimensions of features), an SVD meta-layer first computes the sample covariance as:
\begin{equation}
    \mP = \mX \mJ \mX^{T}, \mJ =\frac{1}{N}(\mI-\frac{1}{N}\mathbf{1}\mathbf{1}^{T}) 
\end{equation}
where $\mJ$ represents the centering matrix, $\mathbf{I}$ denotes the identity matrix, and $\mathbf{1}$ is a column vector whose values are all ones, respectively. The covariance is always positive semi-definite (PSD) and does not have any negative eigenvalues. Afterward, the eigendecomposition is performed using the SVD:
\begin{equation}
    \mP=\mU\mathbf{\Lambda}\mU^{T},\ \mathbf{\Lambda}=\rm{diag}(\lambda_{1},\dots,\lambda_{d})
    \label{SVD}
\end{equation}
where $\mathbf{U}$ is the orthogonal eigenvector matrix, ${\rm diag}(\cdot)$ denotes transforming a vector to a diagonal matrix, and $\mathbf{\Lambda}$ is the diagonal matrix in which the eigenvalues are sorted in a non-increasing order \emph{i.e.}, $\lambda_i {\geq} \lambda_{i+1}$. Then depending on the application, the matrix square root or the inverse square root is calculated as:
\begin{equation}
\begin{gathered}
    \mathbf{Q}\triangleq\mathbf{P}^{\frac{1}{2}}=\mathbf{U}\mathbf{\Lambda}^{\frac{1}{2}} \mathbf{U}^{T}, \mathbf{\Lambda}^{\frac{1}{2}}={\rm diag}(\lambda_{1}^{\frac{1}{2}},\dots,\lambda_{d}^{\frac{1}{2}}) \\ 
    \mathbf{S}\triangleq\mathbf{P}^{-\frac{1}{2}}=\mathbf{U}\mathbf{\Lambda}^{-\frac{1}{2}} \mathbf{U}^{T}, \mathbf{\Lambda}^{-\frac{1}{2}}={\rm diag}(\lambda_{1}^{-\frac{1}{2}},\dots,\lambda_{d}^{-\frac{1}{2}})
\end{gathered}
\end{equation}
The matrix square root $\mQ$ is often used in GCP-related tasks~\cite{li2017second,xie2021so,song2021approximate}, while the application of decorrelated BN~\cite{huang2018decorrelated,siarohin2018whitening} widely applies the inverse square root $\mS$. In certain applications such as WCT, both $\mQ$ and $\mS$ are required. 


\subsection{Backward Pass} 

Let $\frac{\partial l}{\partial\mQ}$ and $\frac{\partial l}{\partial\mS}$ denote the partial derivative of the loss $l$ w.r.t to the matrix square root $\mQ$ and the inverse square root $\mS$, respectively. Then the gradient passed to the eigenvector is computed as:
\begin{equation}
    \frac{\partial l}{\partial \mathbf{U}}\Big|_{\mQ}=(\frac{\partial l}{\partial \mathbf{Q}} + (\frac{\partial l}{\partial \mathbf{Q}})^{T})\mathbf{U}\mathbf{\Lambda}^{\frac{1}{2}},\ 
    \frac{\partial l}{\partial \mathbf{U}}\Big|_{\mS}=(\frac{\partial l}{\partial \mathbf{S}} + (\frac{\partial l}{\partial \mathbf{S}})^{T})\mathbf{U}\mathbf{\Lambda}^{-\frac{1}{2}}
    \label{vec_de}
\end{equation}
Notice that the gradient equations for $\mQ$ and $\mS$ are different. For the eigenvalue, the gradient is calculated as:
\begin{equation}
\begin{gathered}
    \frac{\partial l}{\partial \mathbf{\Lambda}}\Big|_{\mQ}=\frac{1}{2}\rm{diag}(\lambda_{1}^{-\frac{1}{2}},\dots,\lambda_{d}^{-\frac{1}{2}})\mathbf{U}^{T} \frac{\partial \it{l}}{\partial \mathbf{Q}} \mathbf{U}, \\
    \frac{\partial l}{\partial \mathbf{\Lambda}}\Big|_{\mS}=-\frac{1}{2}\rm{diag}(\lambda_{1}^{-\frac{3}{2}},\dots,\lambda_{d}^{-\frac{3}{2}})\mathbf{U}^{T} \frac{\partial \it{l}}{\partial \mathbf{S}} \mathbf{U}
\end{gathered}
\end{equation}
Subsequently, the derivative of the SVD step can be calculated as:
\begin{equation}
    \frac{\partial l}{\partial \mathbf{P}}=\mathbf{U}( (\mathbf{K}^{T}\circ(\mathbf{U}^{T}\frac{\partial l}{\partial \mathbf{U}}))+ (\frac{\partial l}{\partial \mathbf{\Lambda}})_{\rm diag})\mathbf{U}^{T}
    \label{COV_de}
\end{equation}
where $\circ$ denotes the matrix Hadamard product, and the matrix $\mathbf{K}$ consists of entries $K_{ij}{=}{1}/{(\lambda_{i}{-}\lambda_{j})}$ if $i{\neq}j$ and $K_{ij}{=}0$ otherwise. This step is the same for both $\mQ$ and $\mS$. Finally, we have the gradient passed to the feature $\mX$ as:
\begin{equation}
    \frac{\partial l}{\partial \mathbf{X}}=(\frac{\partial l}{\partial \mathbf{P}}+(\frac{\partial l}{\partial \mathbf{P}})^{T})\mathbf{X}\mJ
    \label{X_de}
\end{equation}
With the above rules, the SVD function can be easily inserted into any neural networks and trained end-to-end as a meta-layer. 


\section{Mathematical Derivation and Proof}

\subsection{Derivation of Nearest Orthogonal Gradient}

The problem of finding the nearest orthogonal gradient can be defined as:
\begin{equation}
    \min_{\mR} ||\frac{\partial l}{\partial \mW}-\mR ||_{\rm F}\ subject\ to\ \mR\mR^{T}=\mI
\end{equation}
To solve this constrained optimization problem, We can construct the following error function:
\begin{equation}
    e(\mR) = Tr\Big((\frac{\partial l}{\partial \mW}-\mR)^{T}(\frac{\partial l}{\partial \mW}-\mR)\Big) + Tr\Big(\mathbf{\Sigma} \mR^{T}\mR -\mI \Big) 
\end{equation}
where $Tr(\cdot)$ is the trace measure, and $\mathbf{\Sigma}$ denotes the symmetric matrix Lagrange multiplier. Setting the derivative to zero leads to:
\begin{equation}
\begin{gathered}
   \frac{d e(\mR)}{d \mR} = -2 (\frac{\partial l}{\partial \mW}-\mR) + 2\mR\mathbf{\Sigma} = 0 \\
   \frac{\partial l}{\partial \mW} = \mR(\mI + \mathbf{\Sigma} ),\  \mR = \frac{\partial l}{\partial \mW}(\mI + \mathbf{\Sigma})^{-1}
   \label{inter_result}
\end{gathered}
\end{equation}
The term $(\mI + \mathbf{\Sigma})$ can be represented using $\frac{\partial l}{\partial \mW}$. Consider the covariance of $\frac{\partial l}{\partial \mW}$:
\begin{equation}
\begin{gathered}
   (\frac{\partial l}{\partial \mW})^{T}\frac{\partial l}{\partial \mW} = (\mI + \mathbf{\Sigma} )^{T}\mR^{T}\mR(\mI + \mathbf{\Sigma} ) = (\mI + \mathbf{\Sigma} )^{T}(\mI + \mathbf{\Sigma} ) \\
   (\mI + \mathbf{\Sigma} ) = \Big((\frac{\partial l}{\partial \mW})^{T}\frac{\partial l}{\partial \mW}\Big)^{\frac{1}{2}}
\end{gathered}
\end{equation}
Substituting the term $(\mI + \mathbf{\Sigma})$ in eq.~\eqref{inter_result} with the above equation leads to the closed-form solution of the nearest orthogonal gradient:
\begin{equation}
    \mR = \frac{\partial l}{\partial \mW} \Big(( \frac{\partial l}{\partial \mW})^{T} \frac{\partial l}{\partial \mW}\Big)^{-\frac{1}{2}}
\end{equation}

\subsection{Derivation of Optimal Learning Rate}
To jointly optimize the updated weight $\mW{-}{\eta}\frac{\partial l}{\partial \mW}$, we need to achieve the following objective:
\begin{equation}
   \min_{\eta} ||(\mW{-}{\eta}\frac{\partial l}{\partial \mW})(\mW{-}{\eta}\frac{\partial l}{\partial \mW})^{T}-\mI||_{\rm F}
\end{equation}
This optimization problem can be more easily solved in the form of vector. Let $\mathbf{w}$, $\mi$, and $\mathbf{l}$ denote the vectorized $\mW$, $\mI$, and $\frac{\partial l}{\partial \mW}$, respectively. Then we construct the error function as:
\begin{equation}
   e(\eta) =\Big( (\mathbf{w}-\eta\mathbf{l})^{T}(\mathbf{w}-\eta\mathbf{l})-\mi\Big)^{T}\Big( (\mathbf{w}-\eta\mathbf{l})^{T}(\mathbf{w}-\eta\mathbf{l})-\mi\Big)
\end{equation}
Expanding the equation leads to:
\begin{equation}
     e(\eta)=(\mathbf{w}^{T}\mathbf{w}-2\eta\mathbf{l}^{T}\mathbf{w}+\eta^{2}\mathbf{l}^{T}\mathbf{l}-\mathbf{i})^{T}(\mathbf{w}^{T}\mathbf{w}-2\eta\mathbf{l}^{T}\mathbf{w}+\eta^{2}\mathbf{l}^{T}\mathbf{l}-\mathbf{i})
\end{equation}
Differentiating $e(\eta)$ w.r.t. $\eta$ yields:
\begin{equation}
\begin{gathered}
   \frac{d e(\eta)}{d \eta} = -4\mw\mw^{T}\ml^{T}\mw+ 4\eta\mw\mw^{T}\ml^{T}\ml\\+ 8\eta\ml^{T}\mw\ml^{T}\mw-12\eta^{2}\ml^{T}\mw\ml^{T}\ml+4\ml\mw^{T}\mi
   +4\eta^{3}\ml\ml^{T} - 4\eta\mi\ml\ml^{T}
\end{gathered}
\end{equation}
Since $\eta$ is typically very small, the higher-order terms (\emph{e.g.,} $\eta^{2}$ and $\eta^{3}$) are sufficiently small such that they can be neglected. After omitting these terms, the derivative becomes:
\begin{equation}
    \frac{d e(\eta)}{d \eta}{\approx}-4\mw\mw^{T}\ml^{T}\mw + 4\eta\mw\mw^{T}\ml^{T}\ml + 8\eta\ml^{T}\mw\ml^{T}\mw +4\ml\mw^{T}\mi - 4\eta\mi\ml\ml^{T}\\
\end{equation}
Setting the derivative to zero leads to the optimal learning rate:
\begin{equation}
    \eta^{\star} \approx \frac{\mw^{T}\mw\ml^{T}\mw-\ml^{T}\mw\mi}{\mw^{T}\mw\ml^{T}\ml+2\ml^{T}\mw\ml^{T}\mw - \ml^{T}\ml\mi}
\end{equation}
Notice that $\mi$ is the vectorization of the identify matrix $\mI$, which means that $\mi$ is very sparse ($\emph{i.e.,}$ lots of zeros) and the impact can be neglected. The optimal learning rate can be further simplified as:
\begin{equation}
    \eta^{\star} \approx \frac{\mw^{T}\mw\ml^{T}\mw}{\mw^{T}\mw\ml^{T}\ml+2\ml^{T}\mw\ml^{T}\mw}
\end{equation}



\subsection{Proof of the learning rate bounds}

\begin{duplicate}
 When both $\mW$ and $\frac{\partial l}{\partial \mW}$ are orthogonal, $\eta^{\star}$ is both upper and lower bounded. The upper bound is $\frac{N^2}{N^2 + 2}$ and the lower bound is $\frac{1}{N^{2}+2}$ where $N$ denotes the row dimension of $\mW$.
 \label{prop:lr_bounds}
\end{duplicate}
\begin{proof}

  Since the vector product is equivalent to the matrix Frobenius inner product, we have the relation: 
  \begin{equation}
       \ml^{T}\mw = \langle\frac{\partial l}{\partial \mW},\mW\rangle_{\rm F}
  \end{equation}
  For a given matrix pair $\mA$ and $\mB$, the Frobenius product $\langle\cdot\rangle_{\rm F}$ is defined as:
  \begin{equation}
      \langle\mA,\mB\rangle_{\rm F}=\sum A_{i,j}B_{i,j}\leq \sigma_{1}(\mA)\sigma_{1}(\mB)+\dots+\sigma_{N}(\mA)\sigma_{N}(\mB)
  \end{equation}
  where $\sigma(\cdot)_{i}$ represents the $i$-th largest eigenvalue, $N$ denotes the matrix size, and the inequality is given by Von Neumann’s trace inequality~\cite{mirsky1975trace,grigorieff1991note}. The equality takes only when $\mA$ and $\mB$ have the same eigenvector. When both $\mW$ and $\frac{\partial l}{\partial \mW}$ are orthogonal, \emph{i.e.,} their eigenvalues are all $1$, we have the following relation:
  \begin{equation}
      \langle\frac{\partial l}{\partial \mW},\frac{\partial l}{\partial \mW}\rangle_{\rm F}=N,\ 
      \langle\frac{\partial l}{\partial \mW},\mW\rangle_{\rm F}\leq N
  \end{equation}
  This directly leads to:
  \begin{equation}
      \langle\frac{\partial l}{\partial \mW},\mW\rangle_{\rm F}\leq\langle\frac{\partial l}{\partial \mW},\frac{\partial l}{\partial \mW}\rangle_{\rm F},\ \ml^{T}\mw \leq \ml^{T}\ml 
  \end{equation}
  Exploiting this inequality, the optimal learning rate has the relation:
  \begin{equation}
  \begin{aligned}
       \eta^{\star} \approx \frac{\mw^{T}\mw\ml^{T}\mw}{\mw^{T}\mw\ml^{T}\ml+2\ml^{T}\mw\ml^{T}\mw}
      \leq \frac{\mw^{T}\mw\ml^{T}\ml}{\mw^{T}\mw\ml^{T}\ml+2\ml^{T}\mw\ml^{T}\mw}
  \end{aligned}
      \label{eq:upper_bound_1}
  \end{equation}
  For $\ml^{T}\mw$, we have the inequality as:
  \begin{equation}
  \begin{aligned}
      \ml^{T}\mw &= \langle\frac{\partial l}{\partial \mW},\mW\rangle_{\rm F}=\sum_{i,j} \frac{\partial l}{\partial \mW}_{i,j}\mW_{i,j}\\
      &\geq \sigma_{min}(\frac{\partial l}{\partial \mW})\sigma_{min}(\mW)=1
  \end{aligned}
      \label{eq:inequality_lw}
  \end{equation}
  Then we have the upper bounded of $\eta^{\star}$ as:
  \begin{equation}
  \begin{aligned}
      \eta^{\star} &\leq \frac{\mw^{T}\mw\ml^{T}\ml}{\mw^{T}\mw\ml^{T}\ml+2\ml^{T}\mw\ml^{T}\mw}\\
      &= \frac{N^2}{N^2+2\ml^{T}\mw\ml^{T}\mw} < \frac{N^2}{N^2 + 2}
  \end{aligned}
  \end{equation}
  For the lower bound, since we also have $\ml^{T}\mw{\leq}\mw^{T}\mw$, $\eta^{\star}$ can be re-written as:
  \begin{equation}
  \begin{aligned}
      \eta^{\star} &\approx \frac{\mw^{T}\mw\ml^{T}\mw}{\mw^{T}\mw\ml^{T}\ml+2\ml^{T}\mw\ml^{T}\mw}\\
      &\geq \frac{\ml^{T}\mw\ml^{T}\mw}{\mw^{T}\mw\ml^{T}\ml+2\ml^{T}\mw\ml^{T}\mw}\\
      &= \frac{1}{\frac{\mw^{T}\mw\ml^{T}\ml}{\ml^{T}\mw\ml^{T}\mw}+2} \\
      &=\frac{1}{\frac{N^{2}}{\ml^{T}\mw\ml^{T}\mw}+2}
  \end{aligned}
  \label{eq:lower_bound_1}
  \end{equation}
  Injecting eq.~\eqref{eq:inequality_lw} into eq.~\eqref{eq:lower_bound_1} leads to the further simplification:
  \begin{equation}
     \eta^{\star} \approx \frac{1}{\frac{N^{2}}{\ml^{T}\mw\ml^{T}\mw}+2} \geq \frac{1}{N^{2}+2}
  \end{equation}
  As indicated above, the optimal learning rate $\eta^{\star}$ has a lower bound of $\frac{1}{N^{2}+2}$.
\end{proof}

\begin{figure*}[t]
    \centering
    \includegraphics[width=0.9\linewidth]{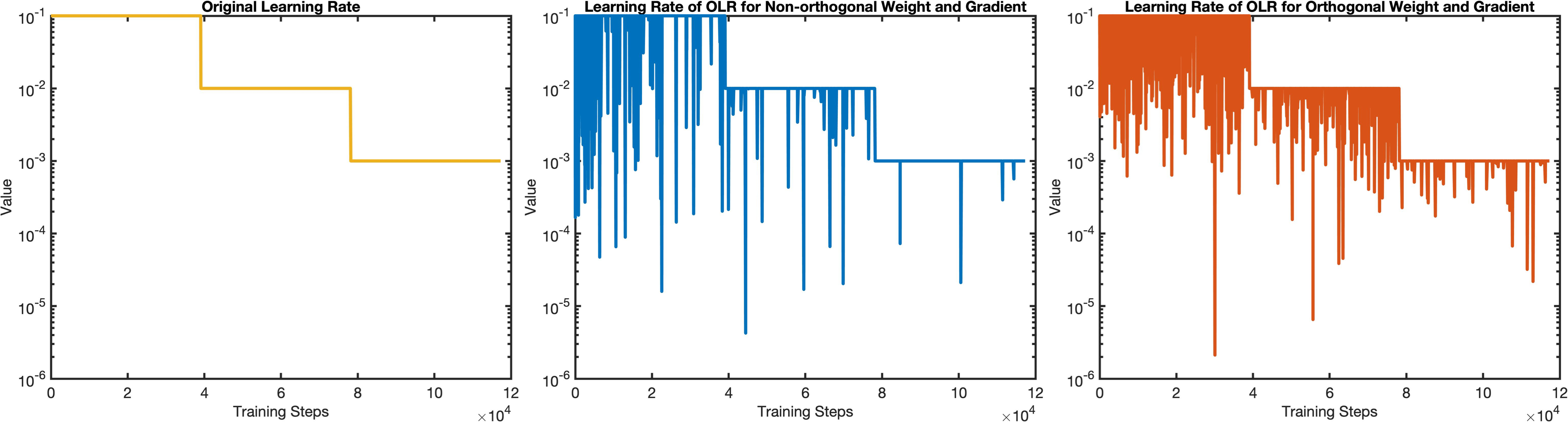}
    \caption{Scheme of learning rate during the training process of decorrelated BN. For the orthogonal weight and gradient, our OLR has a much higher probability of occurrence and can enforce a stronger orthogonality constraint.}
    \label{fig:olr_value}
\end{figure*}

\section{Detailed Experimental Settings}

In this section, we introduce the implementation details and experimental settings.

\subsection{Covariance Conditioning}

\subsubsection{Decorrelated Batch Normalization}

The training lasts $350$ epochs and the learning rate is initialized with $0.1$. The SGD optimizer is used with momentum $0.9$ and weight decay $5e{-}4$. We decrease the learning rate by $10$ every $100$ epochs. The batch size is set to $128$. We use the technique proposed in~\cite{song2021approximate} to compute the stable SVD gradient. The Pre-SVD layer in this experiment is the $3{\times}3$ convolution layer.

\subsubsection{Global Covariance Pooling}

The training process lasts $60$ epochs and the learning rate is initialize with $0.1$. We decrease the learning rate by $10$ at epoch $30$ and epoch $45$. The SGD optimizer is used with momentum $0.9$ and weight decay $1e{-}4$. The model weights are randomly initialized and the batch size is set to $256$. The images are first resized to $256{\times}256$ and then randomly cropped to $224{\times}224$ before being passed to the model. The data augmentation of randomly horizontal flip is used. We use the technique proposed in~\cite{song2021approximate} to compute the stable SVD gradient. The Pre-SVD layer denotes the convolution transform of the previous layer.

\subsection{Latent Disentanglement}



\subsubsection{EigenGAN}

The input image is resize to $128{\times}128$ for AnimeFace~\cite{chao2019/online} and to $256{\times}256$ for FFHQ~\cite{kazemi2014one}. We set the batch size to $128$, and the training process lasts $500,000$ steps. The subspace dimension of each layer is set to $6$, \emph{i.e.,} each layer has $6$ interpretable directions. All the orthogonality techniques are enforced on the projection matrix $\mathbf{U}_{i}$ at each layer. 

\subsubsection{Vanilla GAN}

For both CelebA~\cite{liu2018large} and LSUN Church~\cite{yu2015lsun}, we resize the input image to the resolution of $128{\times}128$. The training lasts $200$ epochs for CelebA and lasts $400$ epochs for LSUN Church. We set the batch size to $128$ and set the latent dimension to $30$.

\section{Occurrence of OLR}

Since our proposed OLR needs manual tuning during the training, it would be interesting to investigate the probability of occurrence in different settings. Fig.~\ref{fig:olr_value} depicts the learning rate schemes of decorrelated BN with ordinary learning rate (\emph{left}), OLR for non-orthogonal weight/gradient (\emph{middle}), and OLR for orthogonal weight/gradient (\emph{right}). As can be seen,
in both settings (orthogonal and non-orthogonal weight/gradient), our OLR occurs with a reasonable probability during the training, which enforces a \emph{related} orthogonality constraint on the weight. When the weight and gradient are non-orthogonal, our OLR mainly occurs at the first training stage where the ordinary learning rate is relative large. For orthogonal gradient and weight, the OLR happens more frequently and consistently occurs throughout all the training stages. This meets our theoretical analysis in Prop.~\ref{prop:lr_bounds}: our OLR suits simultaneously orthogonal weight and gradient.




\end{document}